%% file: tpami.tex
\documentclass[10pt,journal,compsoc]{IEEEtran}
\ifCLASSOPTIONcompsoc
  \usepackage[nocompress]{cite}
\else
  \usepackage{cite}
\fi

\usepackage{graphicx}
\usepackage{amsmath}
\usepackage{amssymb}
\usepackage{wrapfig}
\usepackage{dsfont} 
\usepackage{subcaption} 
\usepackage[letterpaper=true,bookmarks=false,colorlinks]{hyperref}

\ifCLASSINFOpdf
\else
\fi
\begin{document}
%
\title{OpenPose: Realtime Multi-Person 2D Pose Estimation using Part Affinity Fields}

\author{Zhe~Cao,~\IEEEmembership{Student~Member,~IEEE,}
        Gines~Hidalgo,~\IEEEmembership{Student~Member,~IEEE,}
        Tomas~Simon,~\IEEEmembership{}
        Shih-En~Wei,~\IEEEmembership{}
        and~Yaser~Sheikh~\IEEEmembership{}
\IEEEcompsocitemizethanks{\IEEEcompsocthanksitem Z. Cao is with the Berkeley Artificial Intelligence Research lab (BAIR), University of California, Berkeley,
CA, 94709.\protect\\
E-mail: zhecao@berkeley.edu
\IEEEcompsocthanksitem G. Hidalgo and Y. Sheikh are with the Robotics Institute, Carnegie Mellon University, Pittsburgh,
PA, 15213.\protect\\
E-mail: see https://gineshidalgo.com and http://cs.cmu.edu/\textasciitilde yaser
\IEEEcompsocthanksitem T. Simon and S. Wei are with Facebook Reality Labs, Pittsburgh, PA, 15213. 
E-mail: tomas.simon@oculus.com and shih-en.wei@oculus.com}
}

%
%

\markboth{IEEE TRANSACTIONS ON PATTERN ANALYSIS AND MACHINE INTELLIGENCE,~Vol.~XXX, No.~XXX, August~YYYY}%
{Shell \MakeLowercase{\textit{et al.}}: Bare Demo of IEEEtran.cls for Computer Society Journals}


\IEEEtitleabstractindextext{%
\begin{abstract}
Realtime multi-person 2D pose estimation is a key component
in enabling
machines to have an understanding of people in images and videos.
In this work, we present a realtime approach to detect the 2D pose of multiple people in an image.
The proposed method
uses a nonparametric representation, which we refer to as Part Affinity Fields (PAFs), to learn to associate body parts with individuals in the image.
This bottom-up system achieves
high accuracy and realtime performance,
regardless
of the number of people in the image. 
In previous work, PAFs and body part location estimation were refined simultaneously across training stages. We demonstrate that a
PAF-only refinement
rather than both PAF and body part location refinement results in a substantial increase in both runtime performance and accuracy. We also present the first combined body and foot keypoint detector, based on an internal annotated foot dataset that we have publicly released. We show that the combined detector not only reduces the inference time compared to running them sequentially, but also
maintains
the accuracy of each component individually.
This work has culminated in the release of OpenPose, the first open-source realtime system for multi-person 2D pose detection, including body, foot, hand, and facial keypoints.
\end{abstract}

}

\maketitle

\IEEEdisplaynontitleabstractindextext

%
\IEEEpeerreviewmaketitle

\input{section/intro}
\input{section/related_work}

\input{section/method}

\input{section/openpose}


\input{section/evaluation}

\section{Conclusion}
Realtime multi-person 2D pose estimation is a critical component in enabling machines to visually understand and interpret humans and their interactions. In this paper, we
present an explicit nonparametric representation of the keypoint association that encodes both position and orientation of human limbs. Second, we design an architecture
that jointly learns part detection and association. Third, we demonstrate that a greedy parsing algorithm is sufficient to produce
high-quality parses of body poses, and preserves efficiency regardless of the number of people. 
Fourth, we prove that PAF refinement is far more important than combined PAF and body part location refinement, leading to a substantial increase in both runtime performance and accuracy.
Fifth, we show that combining body and foot estimation into a single model boosts the accuracy of each component individually and reduces the inference time of running them sequentially. We have created a foot keypoint dataset consisting of 15K foot keypoint instances, which we will publicly release. Finally, we have 
open-sourced this work as
OpenPose~\cite{hidalgo_cao_simon_wei_joo_sheikh_2017}, the first realtime system for body, foot, hand, and facial keypoint detection.
The library is being widely used today for many research topics involving human analysis, such as human re-identification, retargeting, and Human-Computer Interaction. In addition, OpenPose has been included in the OpenCV library~\cite{opencv_library}.


%



\ifCLASSOPTIONcompsoc
  \section*{Acknowledgments}
\else
  \section*{Acknowledgment}
\fi

We acknowledge the effort from the authors of the MPII and COCO human pose datasets. These datasets make 2D human pose estimation in the wild possible.
This research was supported by the Intelligence Advanced Research Projects Activity (IARPA) via Department of Interior / Interior Business Center (DOI/IBC) contract number D17PC00340.

\ifCLASSOPTIONcaptionsoff
  \newpage
\fi



%




\bibliographystyle{IEEEtran}
\bibliography{egbib}

%

\newpage

\vspace{-15pt}
\begin{IEEEbiography}[{\includegraphics[width=1in,height=1.25in,clip,keepaspectratio]{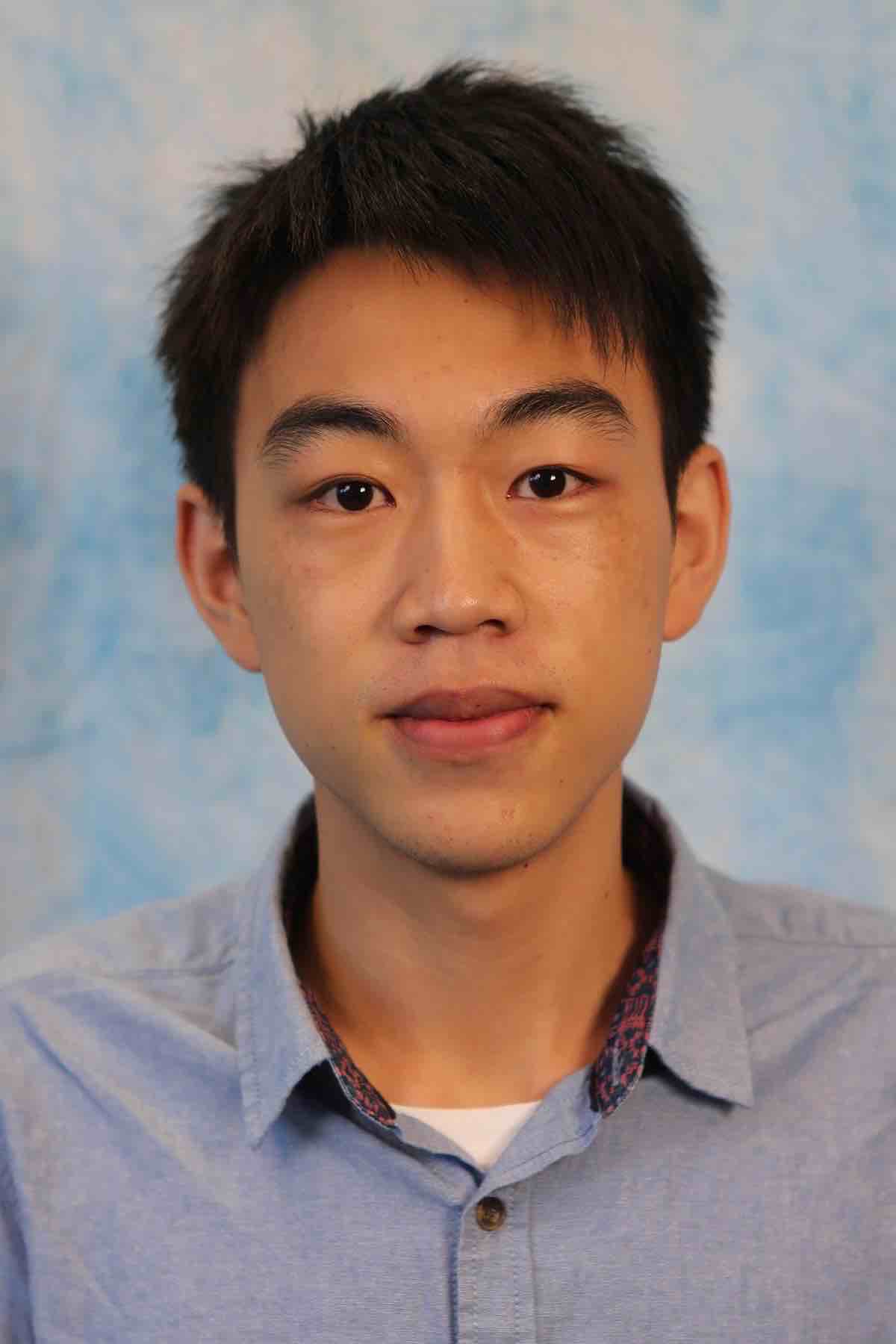}}]{Zhe Cao} is a Ph.D. student in computer vision at University of California, Berkeley, advised by Dr. Jitendra Malik. He received his M.S. in Robotics in 2017 from Carnegie Mellon University, advised by Dr. Yaser Sheikh. He earned his B.S. in Computer Science in 2015 from  Wuhan University, China. His research interests are mainly in computer vision and deep learning. 
\end{IEEEbiography}

\vspace{-30pt}
\begin{IEEEbiography}[{\includegraphics[width=1in,height=1.25in,clip,keepaspectratio]{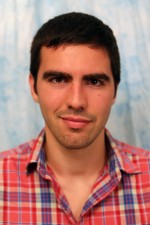}}]{Gines Hidalgo}
received a B.S. degree in Telecommunications 
from Universidad Politecnica de Cartagena, 
Spain, in 2014. He is currently working toward a M.S. degree in robotics at Carnegie Mellon University, 
advised by 
Yaser Sheikh.
Since 2014, he has been a Research Associate in the Robotics Institute at Carnegie Mellon University. His research interests include human pose estimation, computationally efficient deep learning, and computer vision.
\end{IEEEbiography}

\vspace{-30pt}
\begin{IEEEbiography}[{\includegraphics[width=1in,height=1.25in,clip,keepaspectratio]{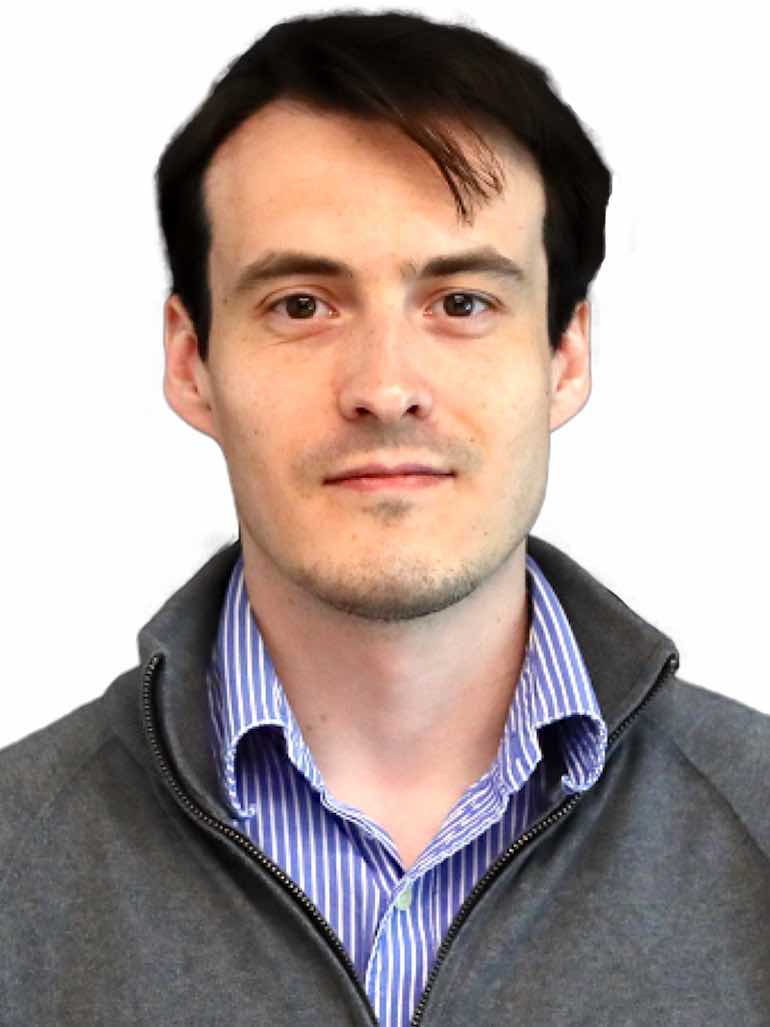}}]{Tomas Simon}
 is a Research Scientist at Facebook Reality Labs. He received his Ph.D. in 2017 from Carnegie Mellon University, advised by Yaser Sheikh and Iain Matthews. He holds a B.S. in Telecommunications from Universidad Politecnica de Valencia and an M.S. in Robotics from Carnegie Mellon University, which he obtained while working at the Human Sensing Lab advised by Fernando De la Torre. His research interests lie mainly in using computer vision and machine learning to model faces and bodies.
\end{IEEEbiography}

\vspace{-30pt}
\begin{IEEEbiography}[{\includegraphics[width=1in,height=1.25in,clip,keepaspectratio]{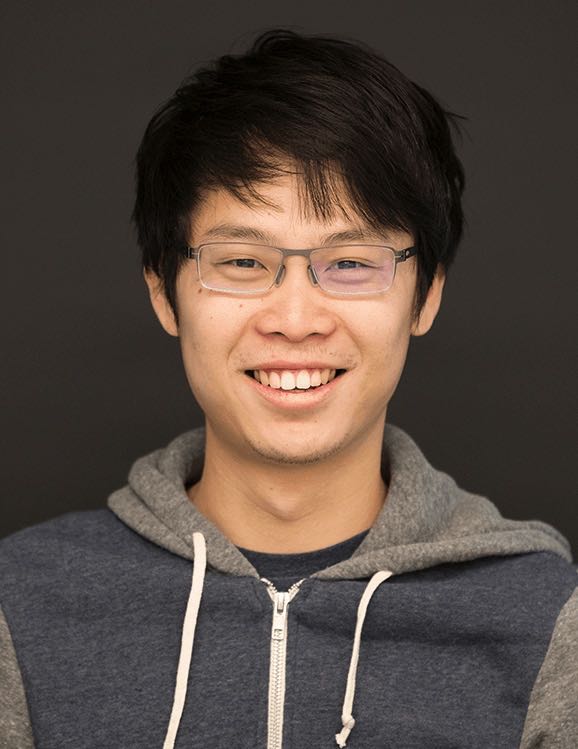}}]{Shih-En Wei}
 is a Research Engineer at Facebook Reality Labs. He received his M.S. in Robotics in 2016 from Carnegie Mellon University, advised by Dr. Yaser Sheikh. He also holds a M.S. in Communication Engineering and B.S. in Electrical Engineering from National Taiwan University, Taipei, Taiwan. His research interests include computer vision and machine learning.
\end{IEEEbiography}

\vspace{-30pt}
\begin{IEEEbiography}[{\includegraphics[width=1in,height=1.25in,clip,keepaspectratio]{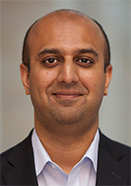}}]{Yaser Sheikh}
is the director of the Facebook Reality Lab in Pittsburgh and is an Associate Professor at the Robotics Institute at Carnegie Mellon University. His research is broadly focused on machine perception of social behavior, spanning computer vision, computer graphics, and machine learning. With colleagues, he has won Popular Science's ``Best of What's New" Award, the Honda Initiation Award (2010), best student paper award at CVPR (2018), best paper awards at WACV (2012), SAP (2012), SCA (2010), ICCV THEMIS (2009), best demo award at ECCV (2016), and placed first in the MSCOCO Keypoint Challenge (2016); he has also received the Hillman Fellowship for Excellence in Computer Science Research (2004). Yaser has served as a senior committee member at leading conferences in computer vision, computer graphics, and robotics including SIGGRAPH (2013, 2014), CVPR (2014, 2015, 2018), ICRA (2014, 2016), ICCP (2011), and served as an Associate Editor of CVIU. His research is sponsored by various government research offices, including NSF and DARPA, and several industrial partners including the Intel Corporation, the Walt Disney Company, Nissan, Honda, Toyota, and the Samsung Group. His research has been featured by various media outlets including The New York Times, The Verge, Popular Science, BBC, MSNBC, New Scientist, slashdot, and WIRED.
\end{IEEEbiography}




\end{document}

%% file: section/intro.tex
\IEEEraisesectionheading{\section{Introduction}\label{sec:intro}}

\IEEEPARstart{I}{n} this paper, we consider a core component in obtaining a  detailed understanding of people in images and videos:
human 2D pose estimation---or the problem of localizing anatomical keypoints or ``parts". Human estimation has largely focused on finding body parts of \emph{individuals}. Inferring the pose of multiple people in images 
presents a unique set of challenges. First, each image may contain an unknown number of people that can appear at any position or scale. Second, interactions between people induce complex spatial interference, due to contact, occlusion,
or
limb articulations, making association of parts difficult. Third, runtime complexity tends to grow with the number of people in the image, making realtime performance a challenge.

\begin{figure}[t!]
    \centering
    \includegraphics[width=1\linewidth]{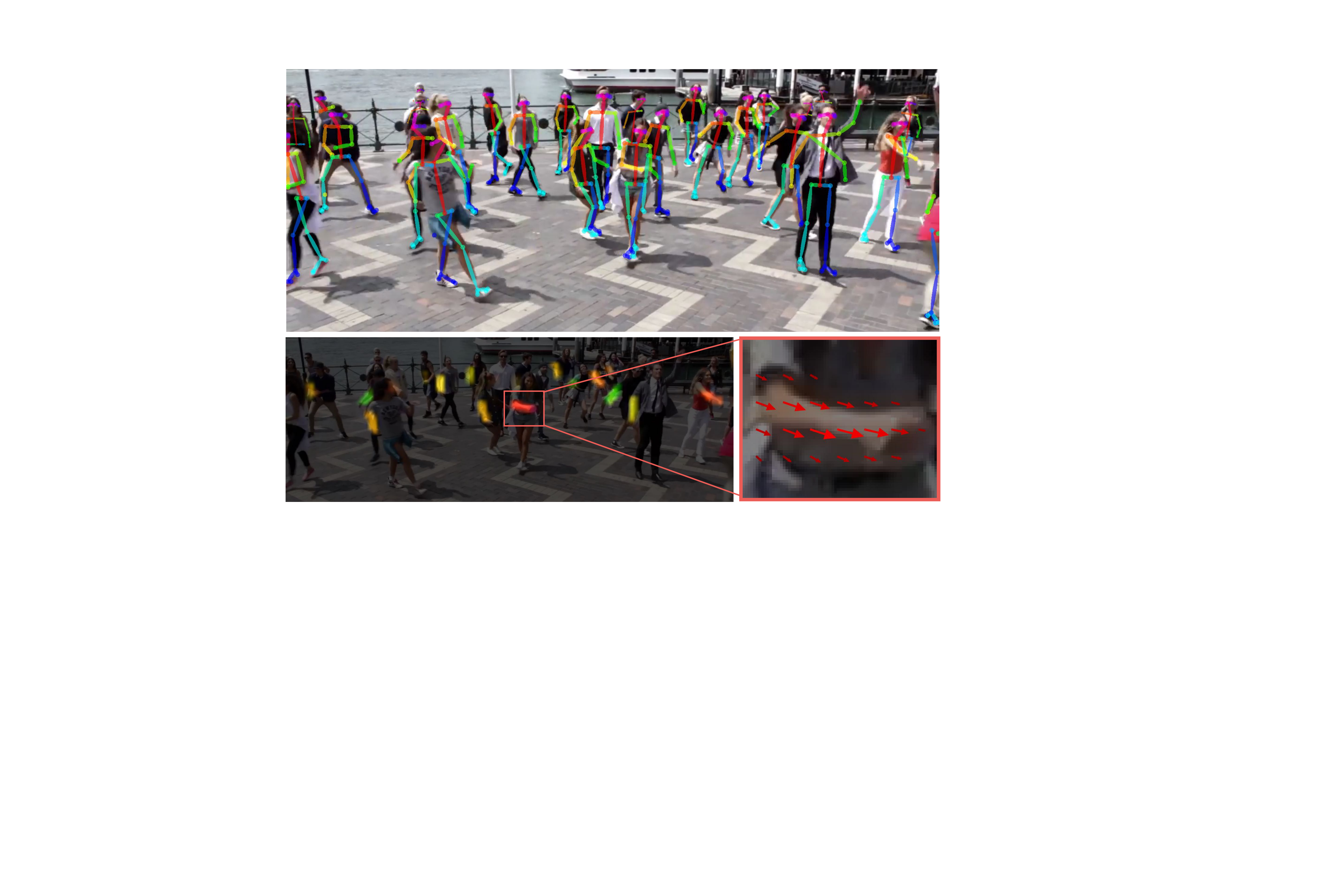} \\
    \vspace{-3pt}
    \caption{\textbf{Top:} Multi-person pose estimation. Body parts belonging to the same person are linked, including foot keypoints (big toes, small toes, and heels). \textbf{Bottom left:} Part Affinity Fields (PAFs) corresponding to the limb connecting right elbow and 
    wrist. The color encodes orientation. \textbf{Bottom right:}
    A 2D vector in each pixel of every PAF encodes the position and orientation of the limbs.}
    \label{fig:teaser}
    \vspace{-12pt}
\end{figure}

\begin{figure*}[t]
    \centering
    \includegraphics[width=0.97\linewidth]{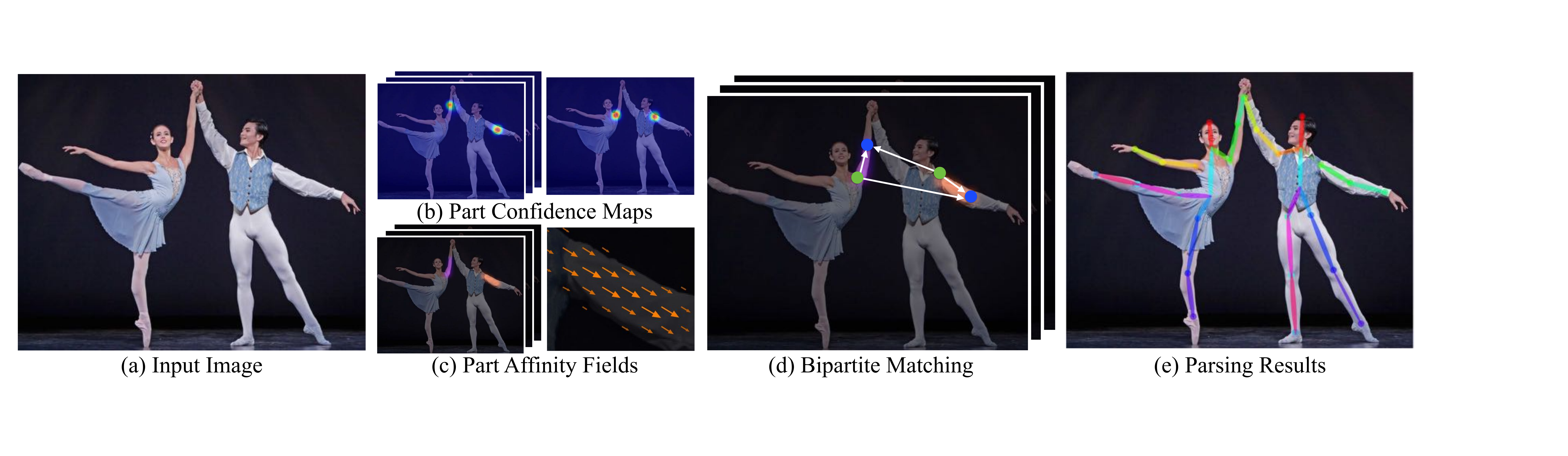} \\
    \vspace{-3pt}
    \caption{Overall pipeline. (a) Our method takes the entire image as the input for a CNN to jointly predict (b) confidence maps for body part detection 
    and (c) PAFs for part association. 
    (d) The parsing step performs a set of bipartite matchings to associate body part candidates. (e) We finally assemble them into full body poses for all people in the image.}
    \label{fig:pipeline}
    \vspace{-12pt}
\end{figure*}

A common approach is to employ a person detector and perform single-person pose estimation for each detection. These top-down approaches directly leverage existing techniques for single-person pose estimation, but suffer from early commitment: if the person detector fails--as it is prone to do when people are in close proximity--there is no recourse to recovery. Furthermore, their runtime
is proportional to the number of people in the image,
for each person detection, a single-person pose estimator is run.
In contrast, bottom-up approaches are attractive as they offer robustness to early commitment and have the potential to decouple runtime complexity from the number of people in the image. Yet, bottom-up approaches do not directly use global contextual cues from other body parts and other people.
Initial
bottom-up methods (\hspace{1sp}\cite{pishchulin2015deepcut,insafutdinov2016deepercut})
did
not retain the gains in efficiency as the final parse
required
costly global inference,
taking
several minutes per image.

In this paper, we present an efficient method for multi-person pose estimation with competitive performance on multiple public benchmarks. We present the first bottom-up representation of association scores via Part Affinity Fields (PAFs), a set of 2D vector fields that encode the location and orientation of limbs over the image domain. We demonstrate that simultaneously inferring these bottom-up representations of detection and association encodes sufficient global context for a greedy parse to achieve high-quality results, at a fraction of the computational cost.

An earlier version of this manuscript appeared in\cite{cao2017realtime}. This version makes several new contributions.
First,
we prove that PAF refinement is crucial for maximizing accuracy, while body part prediction refinement is not that important.
We increase the network depth but remove the body part refinement stages (Sections~\ref{sec:network_arch} and~\ref{sec:simultaneous_det_and_ass}).
This refined network
increases both speed and accuracy by approximately 200\% and 7\%, respectively
(Sections~\ref{sec:coco} and~\ref{sec:runtime}).
Second,
we present an annotated foot dataset\footnote{Dataset webpage:~\url{https://cmu-perceptual-computing-lab.github.io/foot_keypoint_dataset/}} with 15K human foot instances that has been publicly released (Section~\ref{sec:foot}), and we show that
a combined model with body and foot keypoints can be trained preserving the speed of the body-only model while maintaining its accuracy
(Section~\ref{sec:results_foot}).
Third, we demonstrate the generality of our method by applying it to the task of vehicle keypoint estimation (Section~\ref{sec:vehicle}). Finally, this work documents the release of OpenPose~\cite{hidalgo_cao_simon_wei_joo_sheikh_2017}. 
This open-source library is the first available realtime system for multi-person 2D pose detection, including body, foot, hand, and facial keypoints
(Section~\ref{sec:openpose}).
We also include a runtime comparison to Mask R-CNN~\cite{he2017mask} and Alpha-Pose\cite{fang2017rmpe}, showing the computational advantage of our bottom-up approach (Section~\ref{sec:runtime}).

%% file: section/related_work.tex
\section{Related Work}
\textbf{Single Person Pose Estimation} The traditional approach to articulated human pose estimation is to perform inference over a combination of local observations on body parts and the spatial dependencies between them.
The spatial model for articulated pose is either based on tree-structured graphical models~\cite{fh2005pictorial,ramanan2005strike,Andriluka2010,Andriluka2009pictorial,pishchulin2013poselet,yang2013articulated,johnson2010clustered}, which parametrically encode the spatial relationship between adjacent parts following a kinematic chain, or non-tree models~\cite{wang2008multiple,sigal2006measure,lan2005beyond,karlinsky2012using,Dantone2013} that augment the tree structure with additional edges to capture occlusion, symmetry, and long-range relationships. 
To obtain reliable local observations of body parts, Convolutional Neural Networks (CNNs) have been widely used, and have significantly boosted the accuracy on body pose estimation~\cite{newell2016stacked, wei2016convolutional, ouyang2014multi,tompson2014efficient,tompson2014joint,Chen_NIPS14,toshev2014deeppose,belagiannis2017recurrent,bulat2016human,chu2017multi,yang2017learning,chen2017adversarial,tang2018deeply,Ke_2018_ECCV}. Tompson et al.~\cite{tompson2014joint} used a deep architecture with a graphical model whose parameters are learned jointly with the network. 
Pfister et al.~\cite{pfister2015flowing} further used CNNs to implicitly capture global spatial dependencies by designing networks with large receptive fields.
The convolutional pose machines architecture proposed by Wei et al.~\cite{wei2016convolutional} used a multi-stage architecture based on a sequential prediction framework~\cite{ramakrishna2014pose}; iteratively incorporating global context to refine part confidence maps and preserving multimodal uncertainty from previous iterations. Intermediate supervisions are enforced at the end of each stage to address the problem of vanishing gradients~\cite{hochreiter2001gradient,glorot2010understanding,bengio1994learning} during training. Newell et al.~\cite{newell2016stacked} also showed intermediate supervisions are beneficial in a stacked hourglass architecture.
However, all of these methods assume a single person, where the location and scale of the person of interest is given.

\noindent \textbf{Multi-Person Pose Estimation} For multi-person pose estimation, most approaches~\cite{pishchulin2012articulated,gkioxari2014using,sun2011articulated,iqbal2016multi,papandreou2017towards,fang2017rmpe,he2017mask,chen2017cascaded,xiao2018simple} have used a top-down strategy that first detects people and then have estimated the pose of each person independently on each detected region. Although this strategy makes the techniques developed for the single person case directly applicable, it not only suffers from early commitment on person detection, but also fails to capture the spatial dependencies across different people that require global inference.
%
Some approaches have started to consider inter-person dependencies.
Eichner et al.~\cite{eichner2010we} extended pictorial structures to take a set of interacting people and depth ordering into account, 
but still required a person detector to initialize detection hypotheses.
%
Pishchulin et al.~\cite{pishchulin2015deepcut} proposed a bottom-up approach that jointly labels part detection candidates and associated them to individual people, with pairwise scores regressed from spatial offsets of detected parts. This approach does not rely on person detections, however, solving the proposed integer linear programming over the fully connected graph is an NP-hard problem and thus 
the average processing time for a single image is on the order of hours. 
Insafutdinov et al.~\cite{insafutdinov2016deepercut} built on~\cite{pishchulin2015deepcut} with a stronger part detectors based on ResNet~\cite{he2015deep} and image-dependent pairwise scores, and vastly improved the runtime with an incremental optimization approach, but the method still takes several minutes per image, 
with a limit of at most 150 part proposals. 
The pairwise representations used in ~\cite{insafutdinov2016deepercut}, which are offset vectors between every pair of body parts, are difficult to regress precisely and thus a separate logistic regression is required to convert the pairwise features into a probability score. 

In earlier work~\cite{cao2017realtime}, we present {\em part affinity fields} (PAFs), a representation consisting of a set of flow fields that encodes unstructured pairwise relationships between body parts of a variable number of people. In contrast to~\cite{pishchulin2015deepcut} and~\cite{insafutdinov2016deepercut}, we can efficiently obtain pairwise scores from PAFs without an additional training step. These scores are sufficient for a
greedy parse
to obtain high-quality results with realtime performance for multi-person estimation. 
Concurrent to this work, Insafutdinov et al.~\cite{insafutdinov2017arttrack} further simplified their body-part relationship graph for faster inference in single-frame model and formulated articulated human tracking as spatio-temporal grouping of part proposals. 
Recenetly, Newell et al.~\cite{newell2017associative} proposed associative embeddings which can be thought as tags representing each keypoint's group. They group keypoints with similar tags into individual people. Papandreou et al.~\cite{papandreou2018personlab} proposed to detect individual keypoints and predict their relative displacements, allowing a greedy decoding process to group keypoints into person instances. 
Kocabas et al.~\cite{kocabas18prn} proposed a Pose Residual Network which receives keypoint and person detections, and then assigns keypoints to detected person bounding boxes. Nie et al.~\cite{nie2018ppn} proposed to partition all keypoint detections using dense regressions from keypoint candidates to centroids of persons in the image.

In this work, we make several extensions to our earlier work~\cite{cao2017realtime}. We prove that PAF refinement is critical and sufficient for high accuracy, removing the body part confidence map refinement while increasing the network depth. This leads to a 
faster and more accurate model.
We also present the first combined body and foot keypoint detector, created from an annotated foot dataset that will be publicly released. We prove that combining both detection approaches not only reduces the inference time compared to running them independently, but also
maintains
their individual accuracy. Finally, we present OpenPose, the first open-source library for real time body, foot, hand, and facial keypoint detection.


%% file: section/method.tex
\section{Method}

 \begin{figure}[t!]
  \centering
  \includegraphics[width=0.97\linewidth]{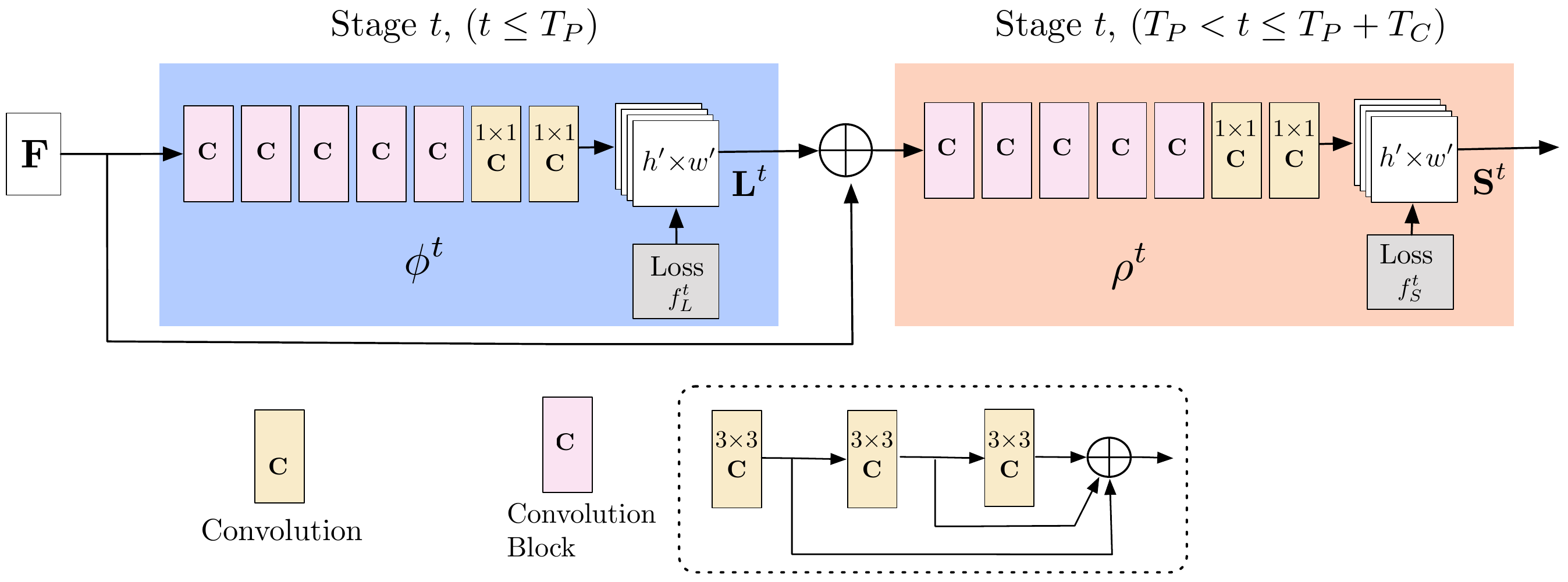}
  \\
  \vspace{-5pt}
   \caption{Architecture of the multi-stage CNN. The first set of stages predicts PAFs $\mathbf{L}^t$, while the last set predicts confidence maps $\mathbf{S}^t$.
   The predictions of each stage and their corresponding image features are concatenated for each subsequent stage.
   Convolutions of kernel size 7 from the original approach~\cite{cao2017realtime} are replaced with 3 layers of convolutions of kernel 3 which are concatenated at their end.}
    \vspace{-10pt}
  \label{fig:arch}
\end{figure}

Fig.~\ref{fig:pipeline} illustrates the overall pipeline of our method. The system takes, as input, a color image of size $w \times h$ (Fig.~\ref{fig:pipeline}a) and produces the 2D locations of anatomical keypoints for each person in the image (Fig.~\ref{fig:pipeline}e). First, a feedforward network predicts a set of 2D confidence maps $\mathbf{S}$ of body part locations (Fig.~\ref{fig:pipeline}b) and a set of 2D vector fields $\mathbf{L}$ of part affinity fields (PAFs), which encode the degree of association between parts (Fig.~\ref{fig:pipeline}c). The set $\mathbf{S}=(\mathbf{S}_1, \mathbf{S}_2, ..., \mathbf{S}_J)$ has $J$ confidence maps, one per part, where $\mathbf{S}_{j}\in \mathds{R}^{w\times h}$, $j \in \{1\ldots J\}$. The set $\mathbf{L} = (\mathbf{L}_1, \mathbf{L}_2, ..., \mathbf{L}_C)$ has $C$ vector fields, one per limb, where $\mathbf{L}_{c}\in \mathds{R}^{w\times h \times 2}$, $c \in \{1\ldots C\}$. We refer to part pairs as limbs for clarity,
but
some pairs are not human limbs (e.g., the face). Each image location in $\mathbf{L}_c$ encodes a 2D vector (Fig.~\ref{fig:teaser}). Finally, the confidence maps and the PAFs are parsed by greedy inference (Fig.~\ref{fig:pipeline}d) to output the 2D keypoints for all people in the image.

\subsection{Network Architecture}\label{sec:network_arch}
Our architecture, shown in Fig.~\ref{fig:arch}, iteratively predicts affinity fields that encode part-to-part association, shown in blue, and detection confidence maps, shown in beige.
The iterative prediction architecture, following
~\cite{wei2016convolutional}, refines the predictions over successive stages, $t\in \{1, \ldots, T\}$, with intermediate supervision at each stage.

The network depth is increased with respect to~\cite{cao2017realtime}.
In the original approach, the network architecture included several 7x7 convolutional layers. In our current model, the receptive field is preserved while the computation is reduced,
by replacing each 7x7 convolutional kernel by 3 consecutive 3x3 kernels.
While the number of operations for the former is $2 \times 7^2 - 1 = 97$, it is only $51$ for the latter. Additionally,
the output of each one of the 3 convolutional kernels is concatenated, following an approach similar to DenseNet~\cite{huang2017densely}.
The number of non-linearity layers is tripled, and the network can keep both lower level and higher level features.
Sections~\ref{sec:coco} and~\ref{sec:runtime} analyze the accuracy and runtime speed improvements, respectively.

\subsection{Simultaneous Detection and Association}\label{sec:simultaneous_det_and_ass}
The image is 
analyzed by a
CNN
(initialized by the first 10 layers of VGG-19~\cite{simonyan2014very} and fine-tuned), generating a set of feature maps $\mathbf{F}$ that is input to the first stage. 
At this stage,
the network produces a set of
part affinity fields (PAFs)
$\mathbf{L}^1 = \phi^{1}(\mathbf{F})$, where $\phi^1$ refers to the CNNs for inference at Stage 1.
In each subsequent stage, the predictions from the previous stage and the original image features $\mathbf{F}$ are concatenated and used to produce refined predictions, 
\begin{eqnarray}
\mathbf{L}^t &=& \phi^{t}(\mathbf{F}, \mathbf{L}^{t-1}),\; \forall 2 \leq t \leq T_P, 
\end{eqnarray}
where $\phi^t$ refers to the CNNs for inference at Stage $t$, and $T_P$ to the number of total PAF stages. After $T_P$ iterations, the process is repeated for the confidence maps detection, starting in the most updated PAF prediction,
\begin{eqnarray}
\mathbf{S}^{T_P} &=& \rho^{t}(\mathbf{F}, \mathbf{L}^{T_P}),\; \forall t = T_P,
\\
\mathbf{S}^t &=& \rho^{t}(\mathbf{F}, \mathbf{L}^{T_P}, \mathbf{S}^{t-1}),\; \forall T_P < t \leq T_P+T_C,
\end{eqnarray}
where $\rho^t$ refers to the CNNs for inference at Stage $t$, and $T_C$ to the number of total confidence map stages.

This approach differs from~\cite{cao2017realtime}, where both the
PAF
and confidence map branches were refined at each stage. Hence, the amount of computation per stage is reduced by half. We empirically observe in Section~\ref{sec:coco} that 
refined affinity field predictions improve the confidence map results, while the opposite does not hold. Intuitively, 
if
we look at the PAF channel output, the body part locations can be guessed. However, if we see a bunch of body parts with no other information, we cannot
parse
them into different people.

 \begin{figure}[t!]
  \centering
  \includegraphics[width=1\linewidth]{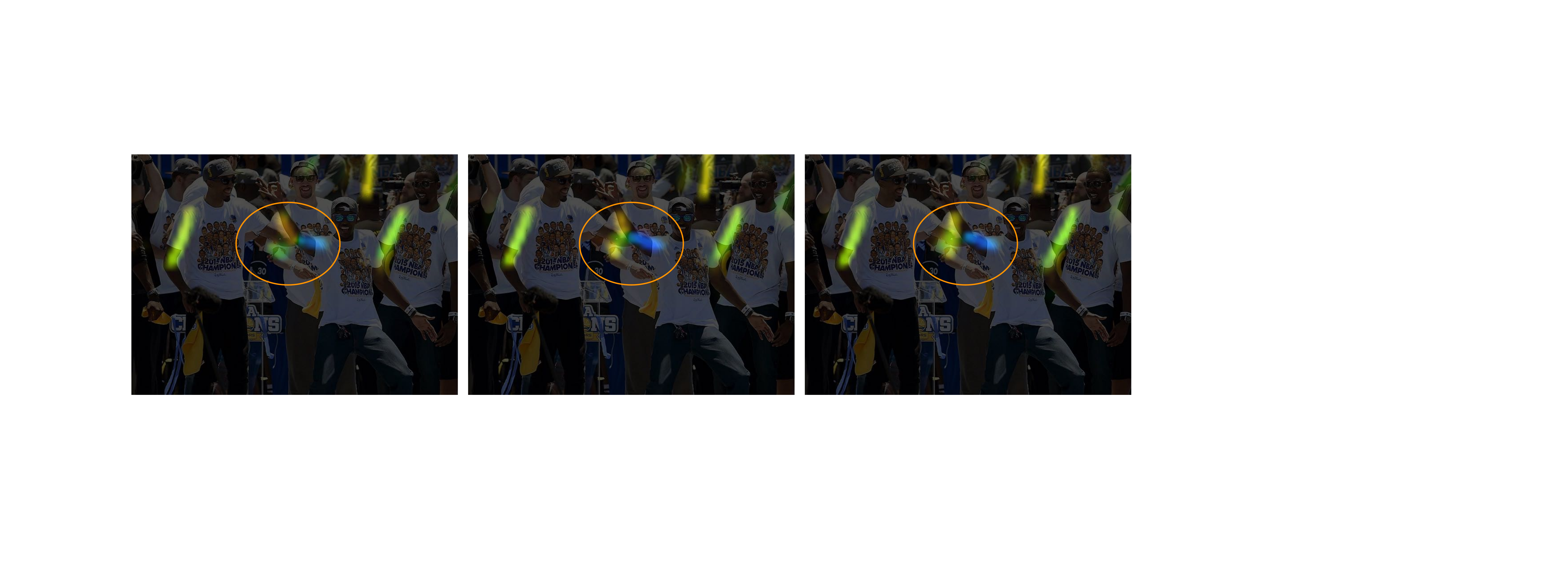}
  \\
  \vspace{-5pt}
  {\footnotesize Stage 1}\hspace{50pt} {\footnotesize Stage 2} \hspace{50pt} {\footnotesize Stage 3}\\
  \caption{\label{fig8} PAFs of right forearm across stages. Although there is confusion between left and right body parts and limbs in early stages, the estimates are increasingly refined through global inference in later stages.}
    \vspace{-10pt}
  \label{fig:stages}
\end{figure}
 
Fig.~\ref{fig8} shows the refinement of the affinity fields across stages.
The confidence map results are predicted on top of the latest and most refined PAF predictions, resulting in a barely noticeable difference across confidence map stages.
To guide the network to iteratively predict PAFs of body parts in the first branch and confidence maps in the second branch, we apply a loss function at the end of each stage. 
We use an $L_2$ loss between the estimated predictions and the groundtruth maps and fields. Here, we weight the loss functions spatially to address a practical issue that some datasets do not completely label all people. Specifically, the loss function of the PAF branch at stage $t_i$ and loss function of the confidence map branch at stage $t_k$ are:
\vspace{-5pt}
 \begin{eqnarray}
     f^{t_i}_\mathbf{L} &=& \sum_{c = 1}^{C} \sum_{\mathbf{p}}  \mathbf{W(\mathbf{p})}\cdot \| \mathbf{L}^{t_i}_{c}(\mathbf{p})  -  \mathbf{L}_{c}^{*}(\mathbf{p}) \|^{2}_{2},\\
     f^{t_k}_\mathbf{S} &=& \sum_{j = 1}^{J} \sum_{\mathbf{p}}  \mathbf{W(\mathbf{p})}\cdot \| \mathbf{S}^{t_k}_{j}(\mathbf{p})  -  \mathbf{S}_{j}^{*}(\mathbf{p}) \|^{2}_{2},
    \label{eqn:localobjective2}
\end{eqnarray}
where $\mathbf{L}_{c}^{*}$ is the groundtruth
PAF,
$\mathbf{S}_{j}^{*}$ is the groundtruth part confidence map, and $\mathbf{W}$ is a binary mask with $\mathbf{W}(\mathbf{p}) = 0$ when the annotation is missing at
the pixel $\mathbf{p}$.
The mask is used to avoid penalizing the true positive predictions during training. 
The intermediate supervision at each stage addresses the vanishing gradient problem by replenishing the gradient periodically~\cite{wei2016convolutional}. The overall objective is
\vspace{-1pt}
\begin{equation}\label{eq:overall}
    f = \sum_{t=1}^{T_P} f_\mathbf{L}^t + \sum_{t=T_P+1}^{T_P+T_C} f_\mathbf{S}^t.
\end{equation}
\vspace{-10pt}

\subsection{Confidence Maps for Part Detection}\label{sec:confidence}
To evaluate $f_\mathbf{S}$ in Eq.~\eqref{eq:overall} during training, we generate the groundtruth confidence maps $\mathbf{S}^*$ from the annotated 2D keypoints. Each confidence map is a 2D representation of the belief that a particular body part
can be located in any given pixel.
Ideally, if a single person appears in the image, a single peak should exist in each confidence map if the corresponding part is visible; if multiple people are in the image, there should be a peak corresponding to each visible part $j$ for each person $k$.

We first generate individual confidence maps $\mathbf{S}_{j,k}^*$ for each person $k$. Let $\mathbf{x}_{j,k}\in\mathds{R}^2$ be the groundtruth position of body part $j$ for person $k$ in the image. The value at location $\mathbf{p}\in\mathds{R}^2$ in $\mathbf{S}_{j,k}^*$ is defined as,
\vspace{-3pt}
\begin{equation}
\mathbf{S}_{j,k}^{*}(\mathbf{p}) = \operatorname{exp}\left(-\frac{||\mathbf{p}-\mathbf{x}_{j,k}||_2^2}{\sigma^2}\right),
\end{equation}
where $\sigma$ controls the spread of the peak. The groundtruth confidence map predicted by the network is an aggregation of the individual confidence maps via a max operator,
\begin{equation}
\mathbf{S}_j^{*}(\mathbf{p}) = \max_k \mathbf{S}_{j,k}^{*}(\mathbf{p}).
\end{equation}

\begin{wrapfigure}{r}{0.43\columnwidth}
\vspace{-0.15in}
\includegraphics[width=1\linewidth]{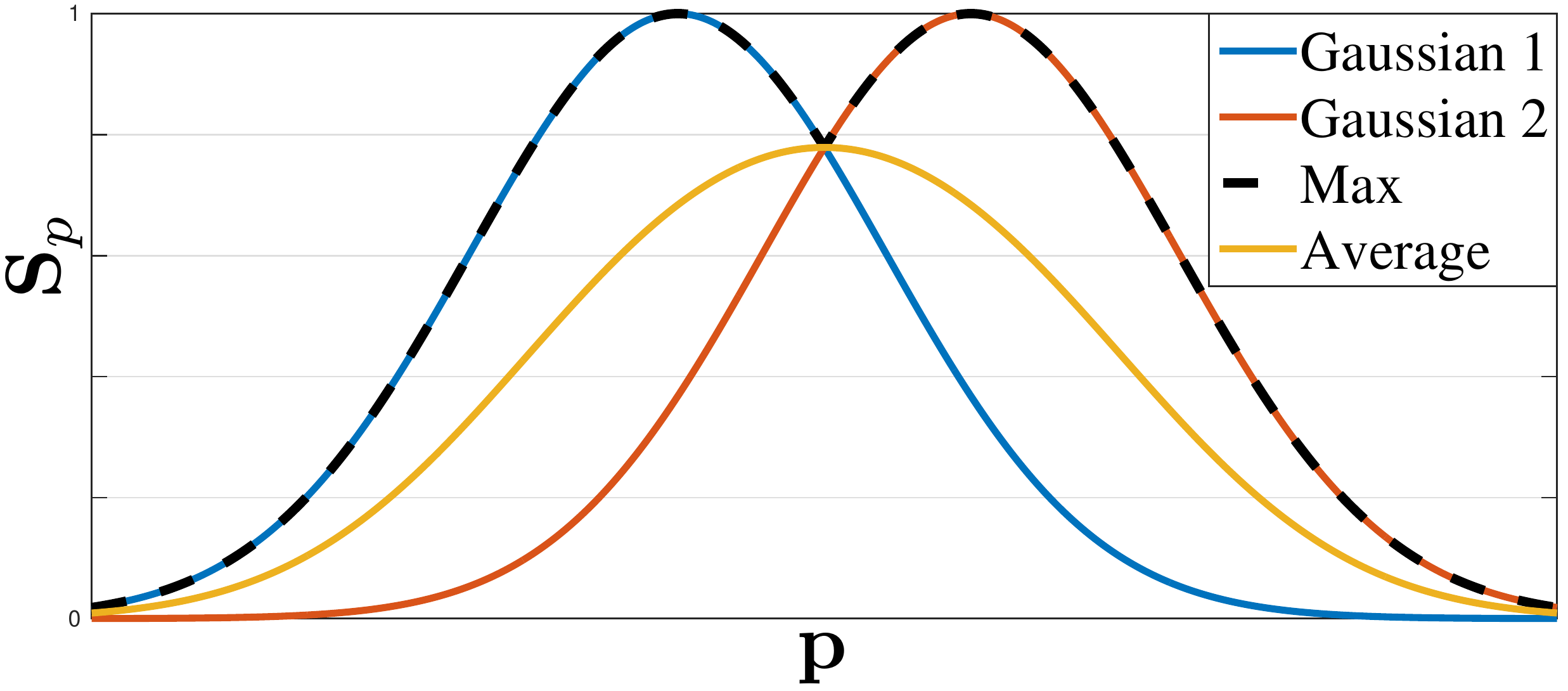}
\vspace{-0.3in}
\end{wrapfigure}
We take the maximum of the confidence maps instead of the average so that the precision of nearby peaks remains distinct, as illustrated in the right figure.
At test time, we predict confidence maps, and obtain body part candidates by performing non-maximum suppression.

\begin{figure}[t!]
    \centering
    \includegraphics[width=0.975\linewidth]{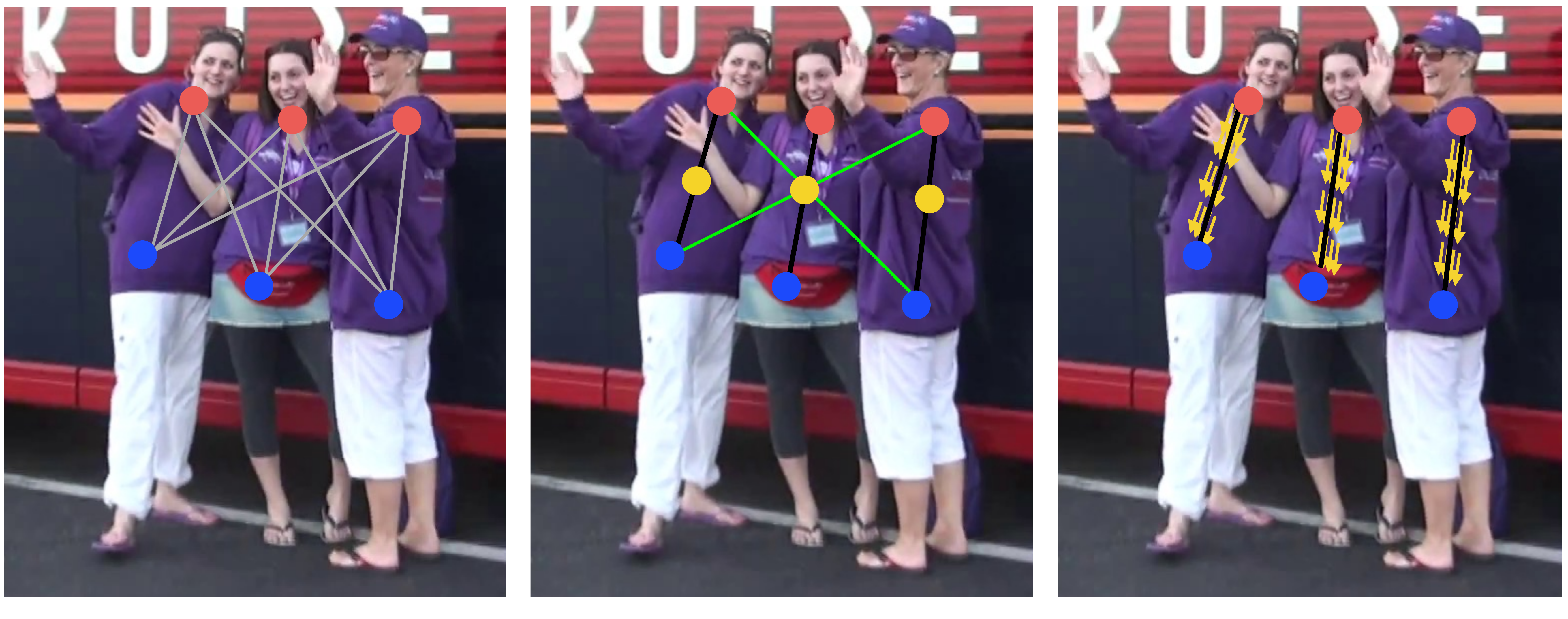}\\
    \vspace{-6pt}
  {\footnotesize (a)}\hspace{71pt}  {\footnotesize (b) }\hspace{71pt} {\footnotesize (c)}\\
    \vspace{-5pt}
    \caption{Part association strategies. (a) The body part detection candidates (red and blue dots) for two body part types and all connection candidates (grey lines). (b) The connection results using the midpoint (yellow dots) representation: correct connections (black lines) and incorrect connections (green lines) that also satisfy the incidence constraint. (c) The results using PAFs (yellow arrows). By encoding position and orientation over the support of the limb, PAFs eliminate false associations.}
    \vspace{-10pt}
    \label{fig:ambiguity}
\end{figure}


\subsection{Part Affinity Fields for Part Association}

Given a set of detected body parts (shown as the red and blue points in Fig.~\ref{fig:ambiguity}a), how do we assemble them to form the full-body poses of an unknown number of people? We need a confidence measure of the association for each pair of body part detections, i.e., that they belong to the same person. One possible way to measure the association is to detect an additional midpoint between each pair of parts on a limb and check for its incidence between candidate part detections, as shown in Fig.~\ref{fig:ambiguity}b. However, when people crowd together---as they are prone to do---these midpoints are likely to support false associations (shown as green lines in Fig.~\ref{fig:ambiguity}b). Such false associations arise due to two limitations in the representation: (1) it encodes only the position, and not the orientation, of each limb; (2) it reduces the region of support of a limb to a single point.

Part Affinity Fields (PAFs) address these limitations. They preserve
both location and orientation information across the region of support of the limb (as shown in Fig.~\ref{fig:ambiguity}c).
Each PAF
is a 2D vector field for each limb, also shown in Fig.~\ref{fig:teaser}d. For each pixel in the area belonging to a particular limb, a 2D vector encodes the direction that points from one part of the limb to the other. Each type of limb has a corresponding
PAF
joining its two associated body parts. 

Consider a single limb shown in the figure below. Let  $\mathbf{x}_{j_1,k}$ and $\mathbf{x}_{j_2,k}$ be the groundtruth positions of body 
\begin{wrapfigure}{r}{0.35\columnwidth}
\vspace{-0.15in}
\includegraphics[width=1\linewidth]{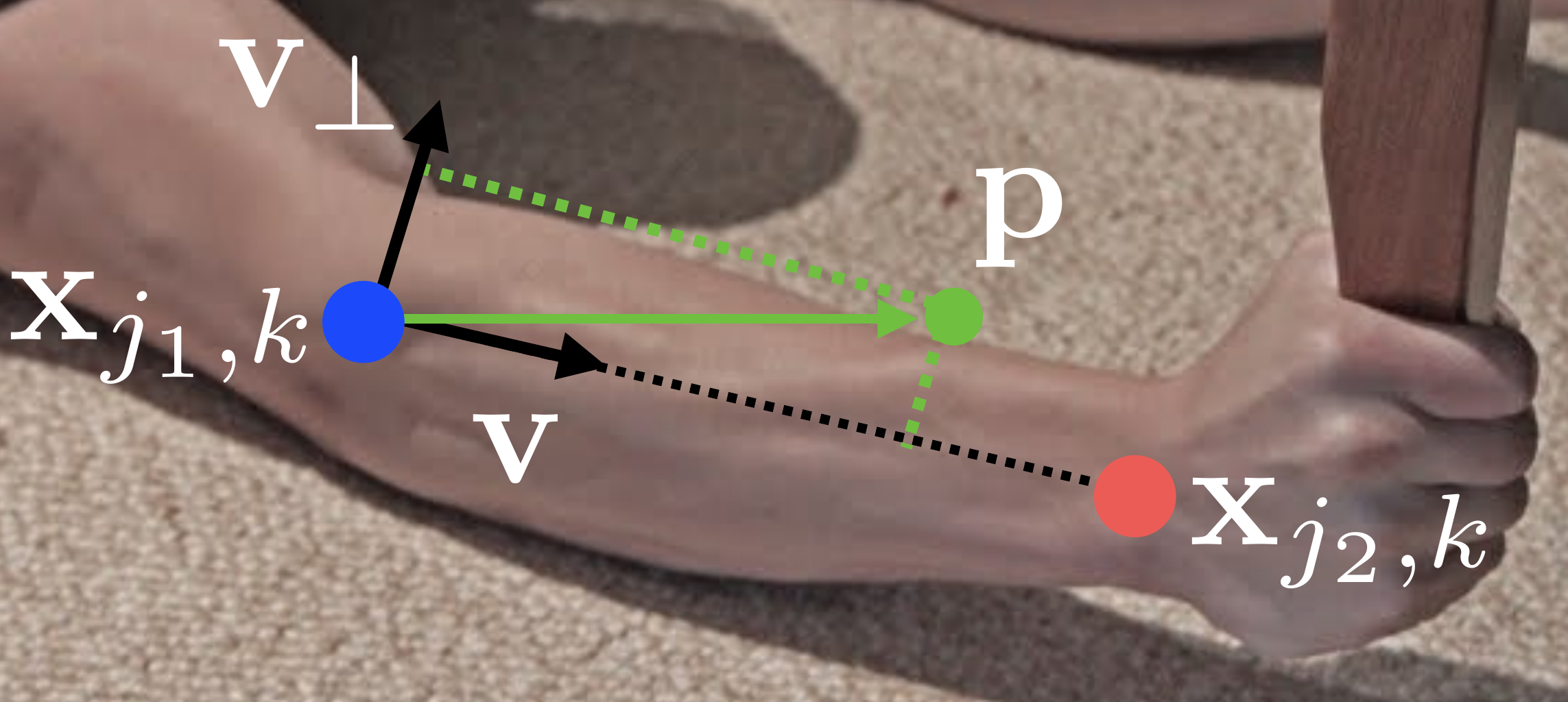}
\vspace{-0.3in}
\end{wrapfigure}
 parts $j_1$ and $j_2$ from the limb $c$ for person $k$ in the image. If a point  $\mathbf{p}$ lies on the limb, the value at $\mathbf{L}_{c,k}^*(\mathbf{p})$ is a unit vector that points from $j_1$ to $j_2$; for all other points, the vector is zero-valued. 

To evaluate $f_\mathbf{L}$ in Eq.~\ref{eq:overall} during training, we define the groundtruth
PAF,
$\mathbf{L}_{c,k}^{*}$, at an image point $\mathbf{p}$ as
\vspace{-3pt}
\begin{equation}
\mathbf{L}_{c,k}^{*}(\mathbf{p}) = \begin{cases}
\mathbf{v} &\textrm{if\ }\mathbf{p}\ \textrm{on limb } c,k\\
\mathbf{0} &\textrm{otherwise.\ }
\end{cases}
\end{equation}
Here, $\mathbf{v}=({\mathbf{x}_{j_2,k}-\mathbf{x}_{j_1,k}})/||\mathbf{x}_{j_2,k}-\mathbf{x}_{j_1,k}||_2$ is the unit vector in the direction of the limb. The set of points on the limb is defined as those within a distance threshold of the line segment, i.e., those points $\mathbf{p}$ for which
\[
0\,{\leq}\,{\mathbf{v}\cdot(\mathbf{p}-\mathbf{x}_{j_1,k})}\,{\leq}\,l_{c,k}\ \textrm{ and }\ |\mathbf{v}_\perp \cdot(\mathbf{p}-\mathbf{x}_{j_1,k})|\,{\leq}\,\sigma_l,
\]
where the limb width $\sigma_l$ is a distance in pixels, the limb length is  $l_{c,k} = ||\mathbf{x}_{j_2,k}-\mathbf{x}_{j_1,k}||_2$, and $\mathbf{v}_\perp$ is a vector perpendicular to $\mathbf{v}$. 

The groundtruth part affinity field averages the affinity fields of all people in the image,
\begin{equation}
\mathbf{L}_{c}^{*}(\mathbf{p}) = \frac{1}{n_c(\mathbf{p})}\sum_{k}\mathbf{L}_{c,k}^{*}(\mathbf{p}),  \vspace{-6pt}
\end{equation}
where $n_c(\mathbf{p})$ is the number of non-zero vectors at point $\mathbf{p}$ across all $k$ people. 

During testing, we measure association between candidate part detections by computing the line integral over the corresponding PAF along the line segment connecting the candidate part locations. In other words, we measure the alignment of the predicted PAF with the candidate limb that would be formed by connecting the detected body parts. Specifically, for two candidate part locations $\mathbf{d}_{j_1}$ and $\mathbf{d}_{j_2}$, we sample the predicted part affinity field, $\mathbf{L}_{c}$ along the line segment to measure the confidence in their association:
\begin{equation}
E = \int_{u=0}^{u=1}  \mathbf{L}_{c}\left( \mathbf{p}(u) \right) \cdot \frac{\mathbf{d}_{j_2} - \mathbf{d}_{j_1}}{||\mathbf{d}_{j_2} - \mathbf{d}_{j_1}||_2}du,
\label{eq:score1}
\end{equation}
where $\mathbf{p}(u)$ interpolates the position of the two body parts $d_{j_1}$ and $d_{j_2}$,
\begin{equation}
 \mathbf{p}(u) = (1-u) \mathbf{d}_{j_1} + u \mathbf{d}_{j_2}.
\end{equation}
\noindent In practice, we approximate the integral by sampling and summing uniformly-spaced values of $u$.

\begin{figure}[t!]
    \centering
    \includegraphics[width=1\linewidth]{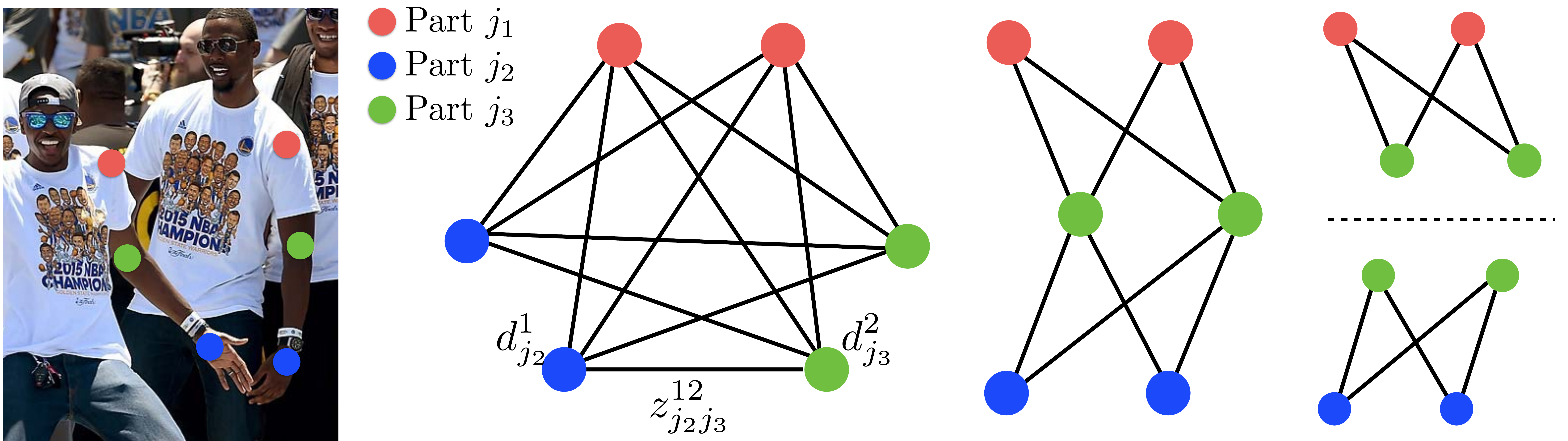}\\
    \vspace{-4pt}
   \hspace{0.2cm} {\footnotesize (a)}\hspace{2.375cm} {\footnotesize (b)}\hspace{2.025cm}  {\footnotesize (c) }\hspace{1.2cm} {\footnotesize (d)}\\
    \vspace{-5pt}
    \caption{Graph matching. (a) Original image with part detections. (b) $K$-partite graph. (c) Tree structure. (d) A set of bipartite graphs.}
    \vspace{-10pt}
    \label{fig:GraphMatching}
\end{figure}

\subsection{Multi-Person Parsing using PAFs} 
We perform non-maximum suppression on the detection confidence maps to obtain a discrete set of part candidate locations. For each part, we may have several candidates, due to multiple people in the image or false positives
(%
Fig.~\ref{fig:GraphMatching}b). These part candidates define a large set of possible limbs. We score each candidate limb using the line integral computation on the PAF, defined in Eq.~\ref{eq:score1}. The problem of finding the optimal parse corresponds to a $K$-dimensional matching problem that is known to be NP-Hard~\cite{west2001introduction}
(%
Fig.~\ref{fig:GraphMatching}c). In this paper, we present a greedy relaxation that consistently produces high-quality matches. We speculate the reason is that the pair-wise association scores implicitly encode global context, due to the large receptive field of the PAF network.

Formally, we first obtain a set of body part detection candidates $\mathcal{D}_\mathcal{J}$ for multiple people, where $\mathcal{D}_\mathcal{J} = \{ \mathbf{d}_j^m : \text{for}\ {j \in \{1 \ldots J\}, m \in \{1 \ldots N_j\}\}}$, where $N_j$ is the number of candidates of part $j$, and $\mathbf{d}_j^m\in\mathds{R}^2$ is the location of the $m$-th detection candidate of body part $j$. These part detection candidates still need to be associated with other parts from the same person---in other words, we need to find the pairs of part detections that are in fact connected limbs. We define a variable $z_{{j_1}{j_2}}^{mn} \in \{0,1\}$ to indicate whether two detection candidates $\mathbf{d}^m_{j_1}$ and $\mathbf{d}^n_{j_2}$ are connected, and the goal is to find the optimal assignment for the set of all possible connections, $\mathcal{Z} = \{ z_{{j_1}{j_2}}^{mn}: \text{for}\ {j_1,j_2 \in \{1 \ldots J\}, m \in \{1 \ldots N_{j_1}\}, n \in \{1 \ldots N_{j_2}\}\}}$.

If we consider a {\em single} pair of parts $j_1$ and $j_2$ (e.g., neck and right hip) for the $c$-th limb, finding the optimal association reduces to a maximum weight bipartite graph matching problem~\cite{west2001introduction}. This case is shown in Fig.~\ref{fig:ambiguity}b. In this graph matching problem, nodes of the graph are the body part detection candidates $\mathcal{D}_{j_1}$ and $\mathcal{D}_{j_2}$, and the edges are all possible connections between pairs of detection candidates. Additionally, each edge is weighted by Eq.~\ref{eq:score1}---the part affinity aggregate. A matching in a bipartite graph is a subset of the edges chosen in such a way that no two edges share a node. Our goal is to find a matching with maximum weight for the chosen edges, 
\begin{eqnarray}
\label{eqn:opt4}
\max_{\mathcal{Z}_c} E_c = \max_{\mathcal{Z}_c} \sum_{m \in \mathcal{D}_{j_1}} \sum_{n \in \mathcal{D}_{j_2} } E_{mn} \cdot z_{{j_1}{j_2}}^{mn},
\end{eqnarray}
\vspace{-5pt}
\begin{eqnarray}
\label{eqn:opt5}
\mathrm{s.t.}&~&\forall {m \in \mathcal{D}_{j_1}}, \sum_{n \in \mathcal{D}_{j_2}} z_{{j_1}{j_2}}^{mn} \leq 1,
\\ \label{eqn:opt6}
&~&\forall {n \in \mathcal{D}_{j_2}}, \sum_{m \in \mathcal{D}_{j_1}} z_{{j_1}{j_2}}^{mn} \leq 1,
\end{eqnarray}

\noindent where $E_c$ is the overall weight of the matching from limb type $c$, $\mathcal{Z}_c$ is the subset of $\mathcal{Z}$ for limb type $c$, and $E_{mn}$ is the part affinity between parts $\mathbf{d}^m_{j_1}$ and $\mathbf{d}^n_{j_2}$ defined in Eq.~\ref{eq:score1}.
Eqs.~\ref{eqn:opt5} and~\ref{eqn:opt6} enforce that no two edges share a node, i.e., no two limbs of the same type (e.g., left forearm) share a part. We can use the Hungarian algorithm~\cite{kuhn1955hungarian} to obtain the optimal matching.

When it comes to finding the full body pose of multiple people, determining $\mathcal{Z}$ is a $K$-dimensional matching problem. This problem is NP-Hard~\cite{west2001introduction} and many relaxations exist. In this work, we add two relaxations to the optimization, specialized to our domain. First, we choose a minimal number of edges to obtain a spanning tree skeleton of human pose rather than using the complete graph, as shown in Fig.~\ref{fig:GraphMatching}c. Second, we further decompose the matching problem into a set of bipartite matching subproblems and determine the matching in adjacent tree nodes independently, as shown in Fig.~\ref{fig:GraphMatching}d. We show detailed comparison results in Section~\ref{sec:mpii}, which demonstrate that minimal greedy inference well-approximates the global solution at a fraction of the computational cost. The reason is that the relationship between adjacent tree nodes is modeled explicitly by PAFs, but internally, the relationship between nonadjacent tree nodes is implicitly modeled by the CNN. This property emerges because the CNN is trained with a large receptive field, and PAFs from non-adjacent tree nodes also influence the predicted PAF.

With these two relaxations, the optimization is decomposed simply as:
\vspace{-7pt}
\begin{equation}
\max_{\mathcal{Z}} E = \sum_{c = 1}^{C} \max_{\mathcal{Z}_c} E_c.
\label{eqn:opt1}
\end{equation}
We therefore obtain the limb connection candidates for each limb type independently using Eqns.~\ref{eqn:opt4}-~\ref{eqn:opt6}. With all limb connection candidates, we can assemble the connections that share the same part detection candidates into full-body poses of multiple people. Our optimization scheme over the tree structure is orders of magnitude faster than the optimization over the fully connected graph~\cite{pishchulin2015deepcut,insafutdinov2016deepercut}.  

\begin{figure}[t]
    \centering
    \begin{subfigure}[t]{0.2375\textwidth}
        \centering
        \includegraphics[width=1\linewidth]{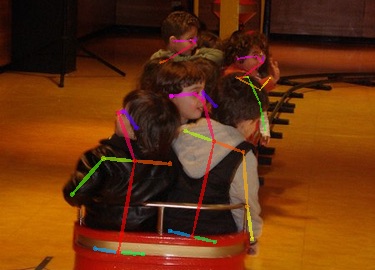}
        \caption{Original Person Parsing}
    \end{subfigure}
    \begin{subfigure}[t]{0.2375\textwidth}
        \centering
        \includegraphics[width=1\linewidth]{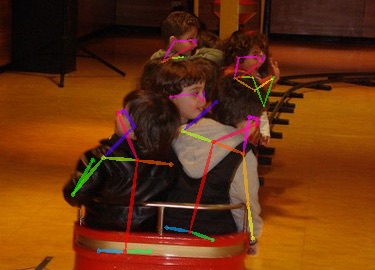}
        \caption{PAF-Redundant Parsing}
    \end{subfigure}
    \vspace{-3pt}
    \caption{Importance of redundant PAF connections. (a) 
    Two different people are wrongly merged due to a wrong neck-nose connection.
    (b) 
    The higher confidence of the right ear-shoulder connection avoids the wrong nose-neck link.}
    \label{fig:redundant_paf_vs_original}
    \vspace{-10pt}
\end{figure}

Our current model also incorporates redundant PAF connections (e.g., between ears and shoulders, wrists and shoulders, etc.). This redundancy particularly improves the accuracy in crowded images, as shown in Fig.~\ref{fig:redundant_paf_vs_original}. To handle these redundant connections, we slightly modify the multi-person parsing algorithm. While the original approach started from a root component, our algorithm sorts all pairwise possible connections by their PAF score. If a connection tries to connect 2 body parts which have already been assigned to different people, the algorithm recognizes that this would contradict a PAF connection with a higher confidence, and the current connection is subsequently ignored.

%% file: section/openpose.tex
\vspace{-2pt}
\section{OpenPose}\label{sec:openpose}

\begin{figure}[t]
    \centering
    \includegraphics[width=1\linewidth]
    {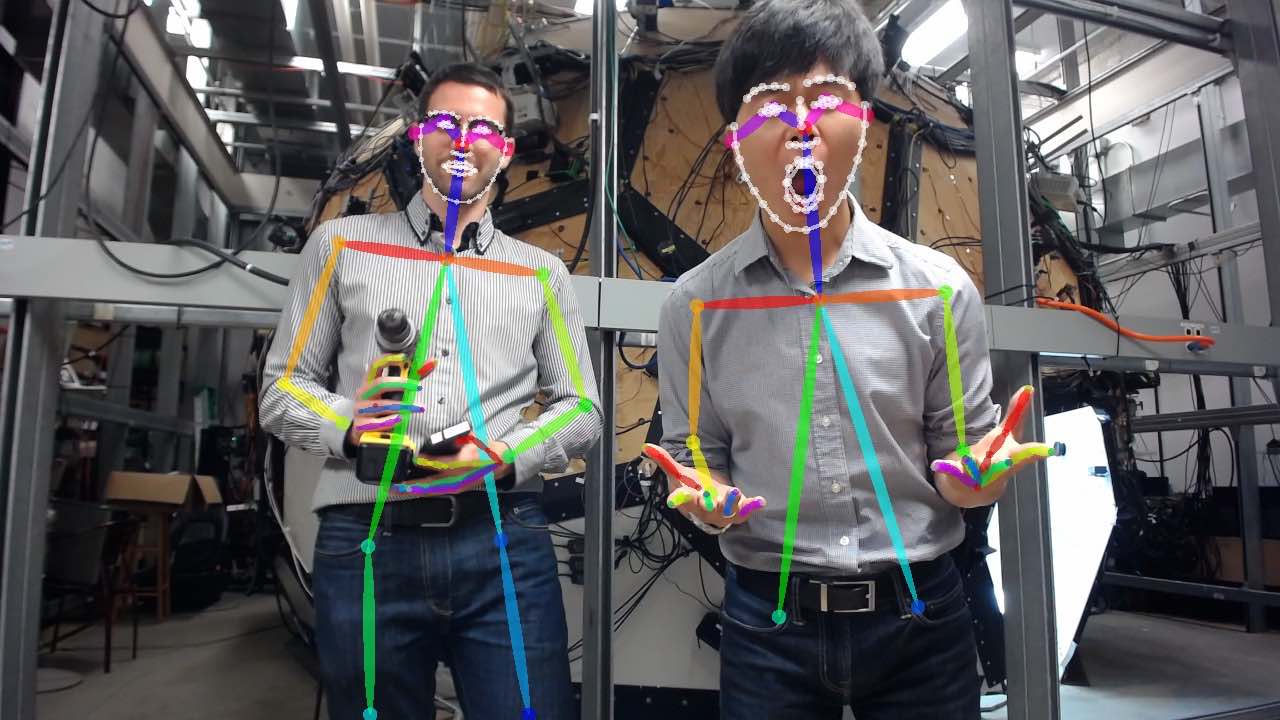}
    \vspace{-15pt}
    \caption{Output of OpenPose, detecting body, foot, hand, and facial keypoints in real-time. OpenPose is robust against occlusions
    including during human-object interaction.
    }
    \label{fig:openpose}
    \vspace{-10pt}
\end{figure}

A growing number of computer vision and machine learning applications require 2D human pose estimation as an input for their systems \cite{qian2017pose, Recycle-GAN, chan2018everybody, gui2018teaching, panteleris2017using, joo2018total, mehta2017vnect}.
To help the research community
boost their work,
we have publicly released
OpenPose~\cite{hidalgo_cao_simon_wei_joo_sheikh_2017}, the first real-time multi-person system to jointly detect human body, foot, hand, and facial keypoints (in total 135 keypoints) on single images. See Fig.~\ref{fig:openpose} for an example of the whole system.

\vspace{-5pt}
\subsection{System}
Available 2D body pose estimation libraries, such as Mask R-CNN~\cite{he2017mask} or Alpha-Pose\cite{fang2017rmpe}, require their users to implement most of the pipeline, their own frame reader (e.g., video, images, or camera streaming), a display to visualize the results, output file generation with the results (e.g., JSON or XML files), etc. In addition, existing facial and body keypoint detectors are not combined, requiring a different library for each purpose.
OpenPose overcome all of these problems.
It can run on different platforms, including Ubuntu, Windows, Mac OSX, and embedded systems (e.g., Nvidia Tegra TX2). It also provides support for different hardware, such as CUDA GPUs, OpenCL GPUs, and CPU-only devices. The user can select an input between images, video, webcam, and IP camera streaming. He can also select whether to display the results or save them on disk, enable or disable each detector (body, foot, face, and hand), enable pixel coordinate normalization, control how many GPUs to use, skip frames for a faster processing, etc.

OpenPose consists of three different blocks: (a) body+foot detection, (b) hand detection~\cite{simon2017hand}, and (c) face detection. The core block is the combined body+foot keypoint detector (Section~\ref{sec:foot}). It can alternatively use the original body-only models~\cite{cao2017realtime} trained on COCO and MPII datasets. Based on the output of the body detector, facial bounding box proposals can roughly be estimated from some body part locations, in particular ears, eyes, nose, and neck. Analogously, the hand bounding box proposals are generated with the arm keypoints. This methodology inherits the problems of top-down approaches discussed in Section~\ref{sec:intro}.
The hand keypoint detector algorithm is explained in further detail in~\cite{simon2017hand}, while the facial keypoint detector has been trained in the same fashion as that of the hand keypoint detector.
The library also includes 3D keypoint pose detection, by performing 3D triangulation with non-linear Levenberg-Marquardt refinement~\cite{marquardt1963algorithm} over the results of multiple synchronized camera views. 

The inference time of OpenPose outperforms all state-of-the-art methods, while preserving high-quality results. It is able to run at about 22 FPS in a machine with a Nvidia GTX 1080 Ti while preserving high accuracy (Section~\ref{sec:runtime}). OpenPose has already been used by the research community for many vision and robotics topics, such as person re-identification~\cite{qian2017pose}, GAN-based video retargeting of human faces~\cite{Recycle-GAN} and bodies~\cite{chan2018everybody}, Human-Computer Interaction~\cite{gui2018teaching}, 3D 
pose estimation~\cite{panteleris2017using}, and 3D human mesh model generation~\cite{joo2018total}. In addition, the OpenCV library~\cite{opencv_library} has included OpenPose and our PAF-based network architecture within its Deep Neural Network (DNN) module.

\begin{figure}[t]
    \centering
    \begin{subfigure}[t]{0.1575\textwidth}
        \centering
        \includegraphics[width=1\linewidth]{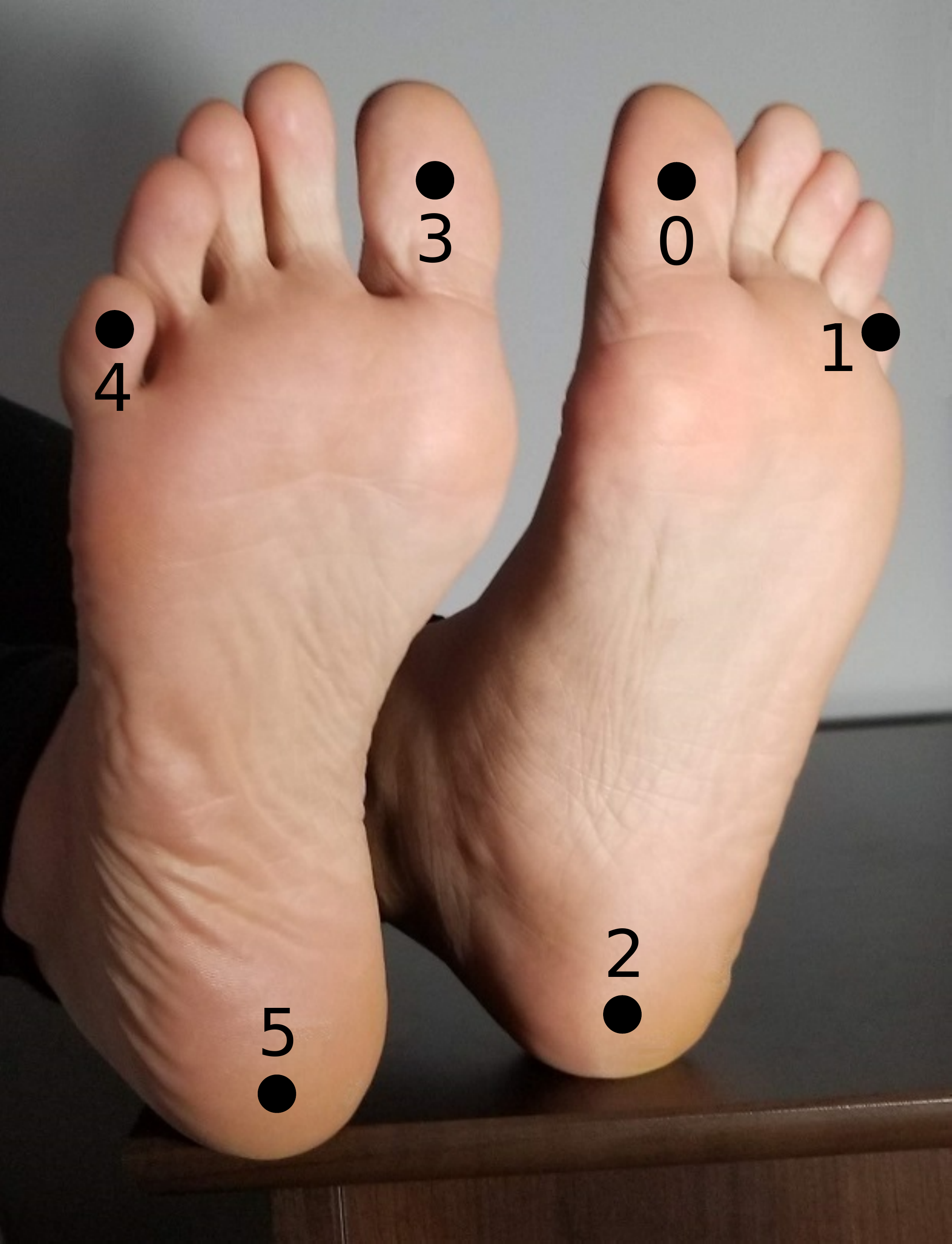}
        \caption{}
    \end{subfigure}
    \begin{subfigure}[t]{0.1575\textwidth}
        \centering
        \includegraphics[width=1\linewidth]{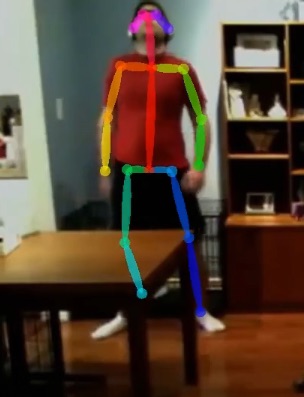}
        \caption{}
    \end{subfigure}
    \begin{subfigure}[t]{0.1575\textwidth}
        \centering
        \includegraphics[width=1\linewidth]{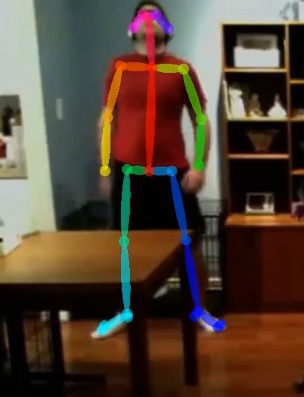}
        \caption{}
    \end{subfigure}
    \vspace{-4pt}
    \caption{Foot keypoint analysis. (a)
    Foot keypoint annotations, consisting of
    big toes, small toes, and heels.
    (b) Body-only 
    model example at which right ankle is not properly estimated.
    (c) Analogous body+foot model example, the foot information helps predict the right ankle location.}
    \label{fig:foot_keypoints}
    \vspace{-10pt}
\end{figure}

\subsection{Extended Foot Keypoint Detection}\label{sec:foot}
Existing human pose datasets (\hspace{1sp}\cite{andriluka20142d, lin2014microsoft}) contain limited body part types. The MPII dataset~\cite{andriluka20142d} annotates ankles, knees, hips, shoulders, elbows, wrists, necks, torsos, and head tops, while COCO~\cite{lin2014microsoft} also includes
some facial keypoints.
For both of these datasets, foot annotations are limited to ankle position only. However, graphics applications such as avatar retargeting or 3D human shape reconstruction~(\hspace{1sp}\cite{joo2018total, loper2015smpl}) require foot keypoints such as big toe and heel.
Without foot information, these approaches suffer from problems such as the candy wrapper effect, floor penetration, and foot skate.
To address these issues,
a small subset of foot instances out of
the COCO dataset 
is labeled using the Clickworker platform~\cite{clickworker}.
It is split up with 14K annotations from the COCO training set and 545 from the validation set.
A total of 6 foot keypoints are labeled
(see Fig.~\ref{fig:foot_keypoints}a). We consider the 3D coordinate of the foot keypoints rather than the surface position. For instance, for the exact toe positions, we label the area between the connection of the nail and skin, and also take depth into consideration by labeling the center of the toe rather than the surface.

\begin{figure}[t]
    \centering
    \begin{subfigure}[t]{0.1575\textwidth}
        \centering
        \includegraphics[width=1\linewidth]{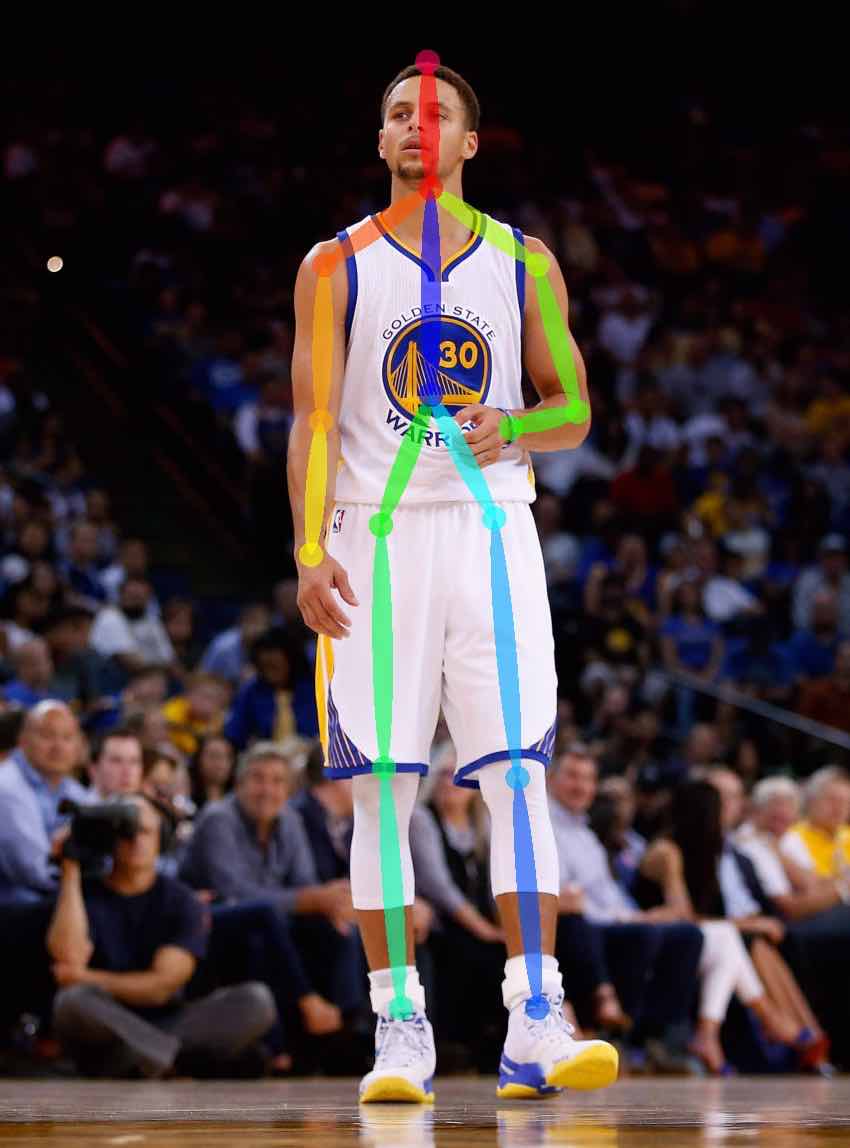}
        \caption{MPII}
    \end{subfigure}
    \begin{subfigure}[t]{0.1575\textwidth}
        \centering
        \includegraphics[width=1\linewidth]{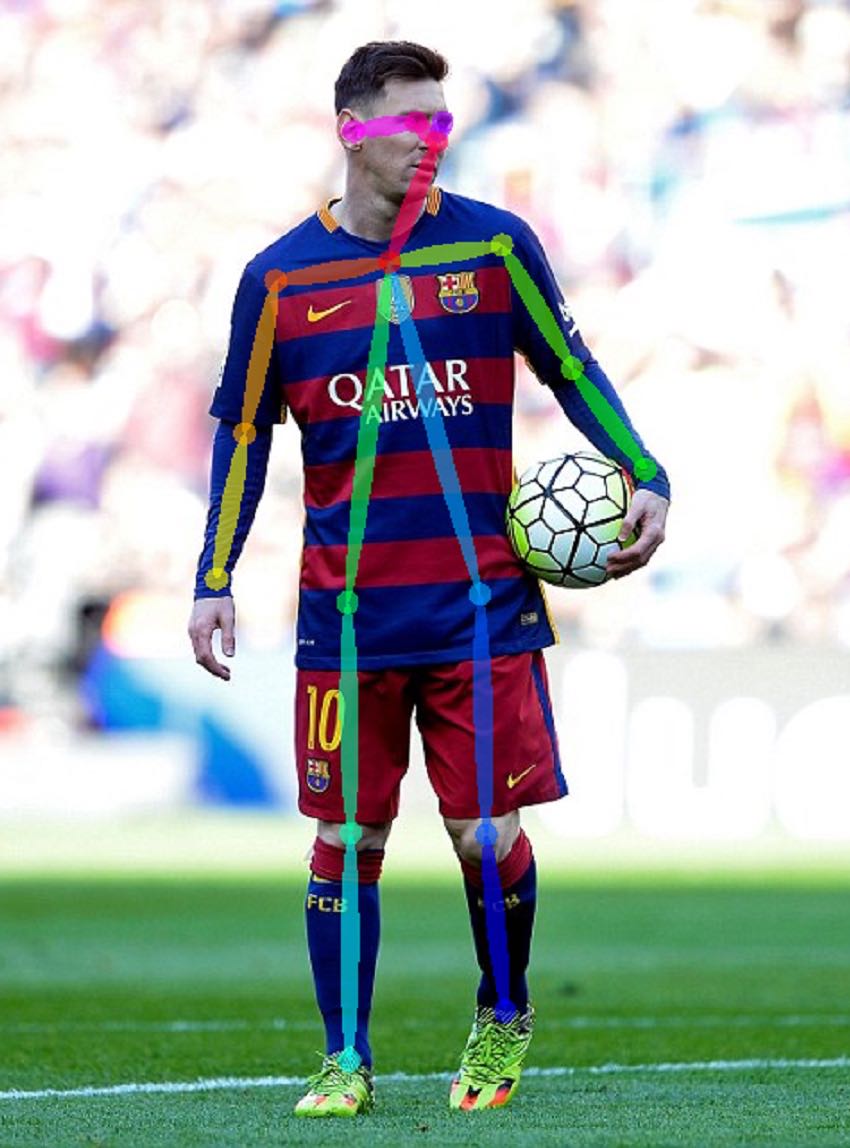}
        \caption{COCO}
    \end{subfigure}
    \begin{subfigure}[t]{0.1575\textwidth}
        \centering
        \includegraphics[width=1\linewidth]{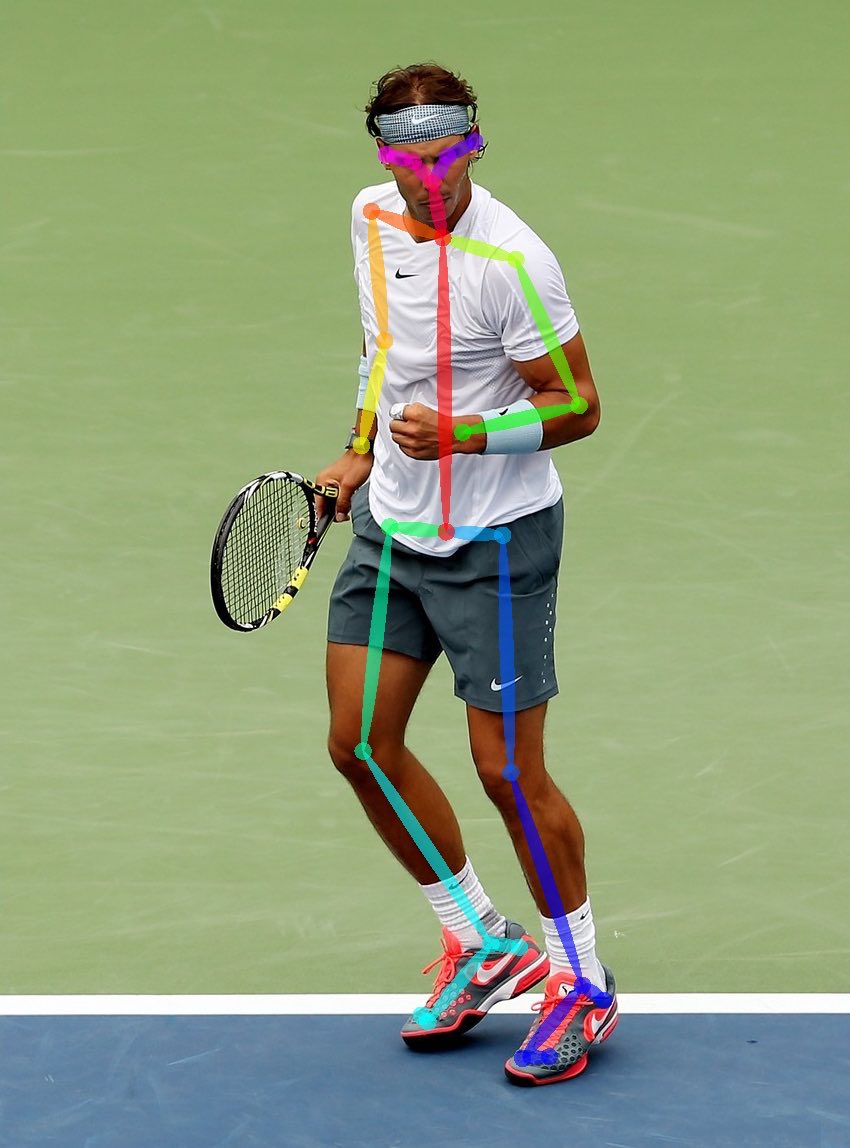}
        \caption{COCO+Foot}
    \end{subfigure}
    \vspace{-3pt}
    \caption{Keypoint annotation configuration for the 3 
    datasets.} 
    \label{fig:datasets}
    \vspace{-10pt}
\end{figure}

Using our dataset, we train a foot keypoint detection algorithm. A n\"{a}ive foot keypoint detector could have been built
by using a body keypoint detector to generate foot bounding box proposals, and then training a foot detector on top of it. 
However, this method suffers from the top-down problems stated in Section~\ref{sec:intro}.
Instead, the same architecture previously described for body estimation is trained to predict both the body and foot locations. Fig.~\ref{fig:datasets} shows the keypoint distribution for the three datasets (COCO, MPII, and COCO+foot). The body+foot model also incorporates an interpolated point between the hips to allow the connection of both legs even when the upper torso is occluded or out of the image.
We find evidence that foot keypoint detection implicitly helps the network to more accurately predict some body keypoints, in particular leg keypoints, such as ankle locations. Fig.~\ref{fig:foot_keypoints}b shows an example where the body-only network was not able to predict ankle location. 
By including foot keypoints during training, while maintaining the same body 
annotations, the algorithm can properly predict the ankle location in Fig.~\ref{fig:foot_keypoints}c. We quantitatively analyze the accuracy
difference
in Section~\ref{sec:results_foot}.

%% file: section/evaluation.tex
\section{Datasets and Evaluations}

We evaluate our method on three benchmarks for multi-person pose estimation: 
(1) MPII human multi-person dataset~\cite{andriluka20142d}, which consists of 3844 training and 1758 testing groups of multiple interacting individuals in highly articulated poses with 14 body parts;
(2) COCO 
keypoint challenge dataset~\cite{lin2014microsoft}, which requires simultaneously detecting people and localizing 17 keypoints (body parts) in each person (including 12 human body parts and 5 facial keypoints);
(3) our foot dataset, which is a subset of 15K annotations out of the COCO keypoint dataset.
These datasets collect images in diverse scenarios that contain many real-world challenges such as crowding, scale variation, occlusion, and contact. Our approach
placed first
at the inaugural COCO 2016 keypoints challenge~\cite{COCOkeypoint}, and significantly
exceeded
the previous state-of-the-art results on the MPII multi-person benchmark. We also provide runtime analysis comparison against Mask R-CNN and Alpha-Pose to quantify the efficiency of the system and analyze the main failure cases.

%
%

%



\subsection{Results on the MPII Multi-Person Dataset}\label{sec:mpii}

For comparison on the MPII dataset, we use the toolkit~\cite{pishchulin2015deepcut} to measure mean Average Precision (mAP) of all body parts following the ``PCKh" metric from~\cite{andriluka20142d}. 
Table~\ref{table:mpi} compares mAP performance between our method and other approaches on the official MPII testing sets. 
We also
compare the average inference/optimization time per image in seconds. For the $288$ images subset, our method outperforms previous state-of-the-art bottom-up methods~\cite{insafutdinov2016deepercut} by $8.5\%$ mAP. Remarkably, our inference time is $6$ orders of magnitude less. We report a more detailed runtime analysis in Section~\ref{sec:runtime}. For the entire MPII testing set, our method without scale search already outperforms previous state-of-the-art methods by a large margin, i.e., $13\%$ absolute increase on mAP. Using a 3 scale search ($\times0.7$, $\times1$ and $\times1.3$) further increases the performance to $75.6\%$ mAP. The mAP comparison with previous bottom-up approaches indicate the effectiveness of our novel feature representation, PAFs, to associate body parts. Based on the tree structure, our greedy parsing method achieves better accuracy than a graphcut optimization formula based on a fully connected graph structure~\cite{pishchulin2015deepcut,insafutdinov2016deepercut}. 

\begin{table}[t]
    \begin{center}
    \resizebox{0.485\textwidth}{!}{
    \begin{tabular}{l | c c c c c c c |c |c}
    \hline
    Method & Hea & Sho & Elb & Wri & Hip & Kne & Ank & \textbf{mAP} & s/image \\
    \hline
    \multicolumn{10}{ c }{Subset of 288 images as in \cite{pishchulin2015deepcut}} \\
    Deepcut~\cite{pishchulin2015deepcut} &73.4 &71.8 &57.9 &39.9 &56.7 &44.0 &32.0 &54.1 &57995\\
    Iqbal et al.~\cite{iqbal2016multi} & 70.0 &65.2 &56.4 &46.1 &52.7 &47.9 &44.5 &54.7 & 10\\
    DeeperCut~\cite{insafutdinov2016deepercut} &87.9 &84.0 &71.9 &63.9 &68.8 &63.8 &58.1 &71.2 &230\\
    Newell et al.~\cite{newell2017associative}&91.5 &87.2	&75.9	&65.4	&72.2	&67.0	&62.1	&74.5& -\\
    ArtTrack~\cite{insafutdinov2017arttrack}& 92.2  & \textbf{91.3}  & 80.8  & 71.4  & \textbf{79.1}  & 72.6 & 67.8 & \textbf{79.3} &\textbf{0.005}\\
    Fang et al.~\cite{fang2017rmpe} &89.3 &88.1 &80.7	&75.5 &73.7	&\textbf{76.7} &\textbf{70.0} &79.1 &-\\
    Ours & \textbf{92.9}	&\textbf{91.3}	&\textbf{82.3}	&72.6	&76.0	&70.9	&66.8	&79.0 &\textbf{0.005}\\
    \hline
    \multicolumn{10}{ c }{Full testing set} \\
    DeeperCut~\cite{insafutdinov2016deepercut} &78.4 &72.5 &60.2 &51.0 &57.2 &52.0 &45.4 &59.5 & 485\\
    Iqbal et al.~\cite{iqbal2016multi} & 58.4 & 53.9 &44.5 &35.0	&42.2 &36.7	&31.1 &43.1 & 10\\
   Levinko et al.~\cite{levinkov2017joint} &89.8	&85.2	&71.8	&59.6	&71.1	&63.0	&53.5	&70.6 &-\\
    ArtTrack~\cite{insafutdinov2017arttrack}& 88.8  & 87.0  & 75.9  & 64.9  & 74.2  & 68.8 & 60.5 & 74.3 &\textbf{0.005}\\
    Fang et al.~\cite{fang2017rmpe} &88.4	&86.5 &78.6	&\textbf{70.4}	&74.4 &\textbf{73.0}	&\textbf{65.8} &76.7 & -\\
    Newell et al.~\cite{newell2017associative}&	\textbf{92.1} &	89.3 &	78.9&	69.8&	76.2&	71.6&	64.7&	77.5& -\\
    Fieraru et al.~\cite{fieraru2018learning} &91.8	&\textbf{89.5}	&\textbf{80.4} &69.6	&\textbf{77.3} &71.7	&65.5	&\textbf{78.0} & -\\
    Ours (one scale) & 89.0  & 84.9  & 74.9  & 64.2  & 71.0  & 65.6 & 58.1 & 72.5 & 0.005\\
    Ours  & 91.2  & 87.6  & 77.7  & 66.8 & 75.4  & 68.9 & 61.7 & 75.6 & \textbf{0.005}\\

    \hline
    \end{tabular}}
    \end{center}
    \vspace{-7pt}
    \caption{Results on the MPII dataset. Top: Comparison results on the testing subset defined in~\cite{pishchulin2015deepcut}. Middle: Comparison results on the whole testing set. Testing without scale search is denoted as ``(one scale)".}
    \label{table:mpi}
\end{table}

\begin{table}[h]
\begin{center}
\resizebox{0.485\textwidth}{!}{
\begin{tabular}{l | c c c c c c c |c |c}
\hline
Method & Hea & Sho & Elb & Wri & Hip & Kne & Ank & \textbf{mAP} & s/image \\
\hline
Fig.~\ref{fig:GraphMatching}b &91.8 &\textbf{90.8} &80.6 &69.5 &78.9 &71.4 &63.8 &78.3 & 362\\ 
Fig.~\ref{fig:GraphMatching}c &92.2 &90.8 &80.2 &69.2 &78.5 &70.7 &62.6 &77.6 & 43\\
Fig.~\ref{fig:GraphMatching}d &92.0 &90.7 &80.0 &69.4 &78.4 &70.1 &62.3 &77.4 & 0.005\\
Fig.~\ref{fig:GraphMatching}d (sep) &\textbf{92.4}  & 90.4  & \textbf{80.9}  & \textbf{70.8} & \textbf{79.5}  & \textbf{73.1} & \textbf{66.5} & \textbf{79.1} & \textbf{0.005} \\
\hline
\end{tabular}}
\end{center}
\vspace{-10pt}
\caption{Comparison of different structures on our custom validation set.}
\vspace{-10pt}
\label{table:fig5}
\end{table}

In Table~\ref{table:fig5}, we show comparison results for the different skeleton structures shown in Fig.~\ref{fig:GraphMatching}. We created a custom validation set consisting of 343 images from the original MPII training set. We train our model based on a fully connected graph, and compare results by selecting all edges (Fig.~\ref{fig:GraphMatching}b, approximately solved by Integer Linear Programming), and minimal tree edges (Fig.~\ref{fig:GraphMatching}c, approximately solved by Integer Linear Programming, and Fig.~\ref{fig:GraphMatching}d, solved by the greedy algorithm presented in this paper).
Both methods yield similar results, demonstrating that it is sufficient to use minimal edges.
We trained our final model to only learn the minimal edges to fully utilize the network capacity, 
denoted as Fig.~\ref{fig:GraphMatching}d (sep). This approach outperforms Fig.~\ref{fig:GraphMatching}c and even Fig.~\ref{fig:GraphMatching}b, while maintaining efficiency.
The fewer number of part association channels (13 edges of a tree vs 91 edges of a graph) needed facilitates the training convergence.

\begin{figure}[t]
\centering
  \includegraphics[width=0.5\linewidth]
  {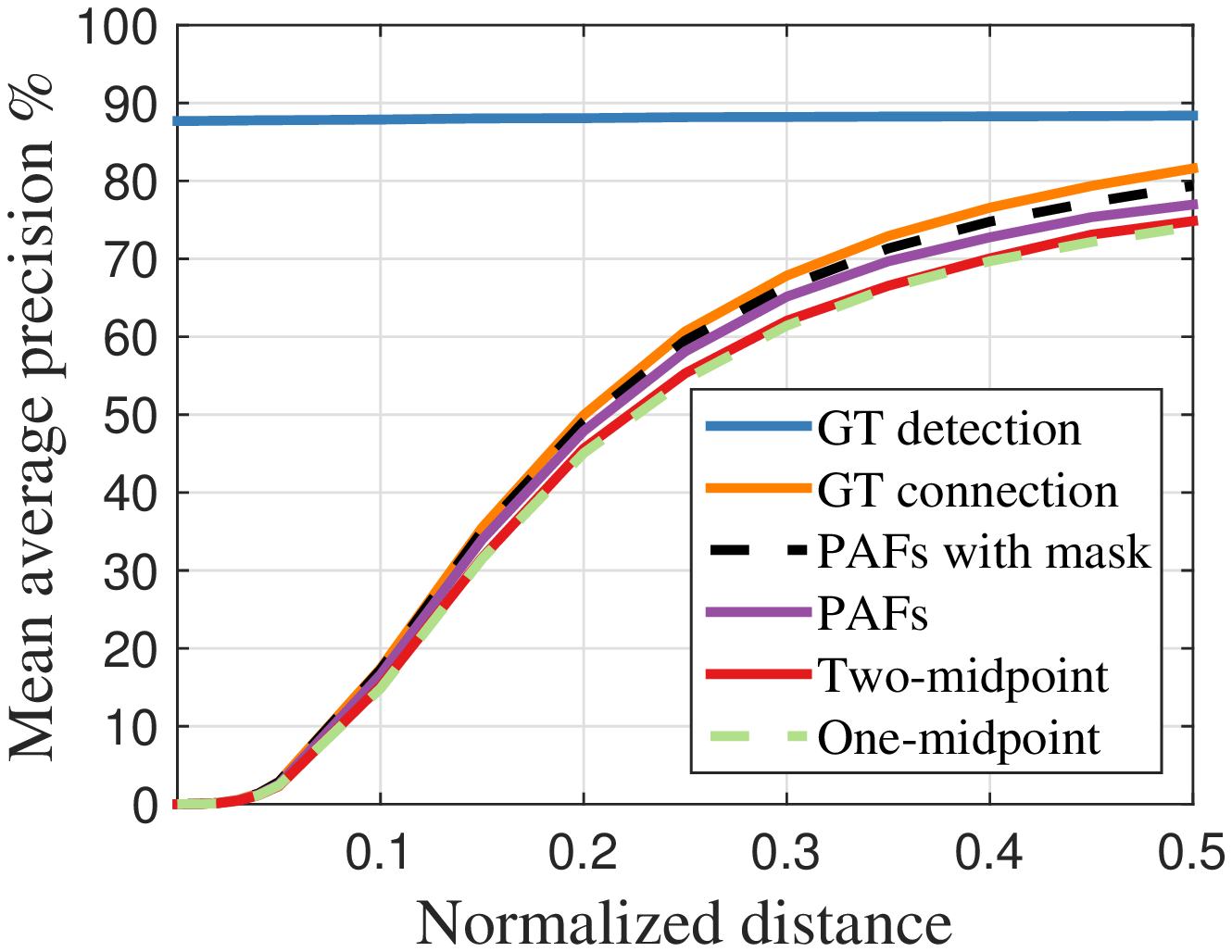}
    \hfill
  \includegraphics[width=0.46\linewidth]
  {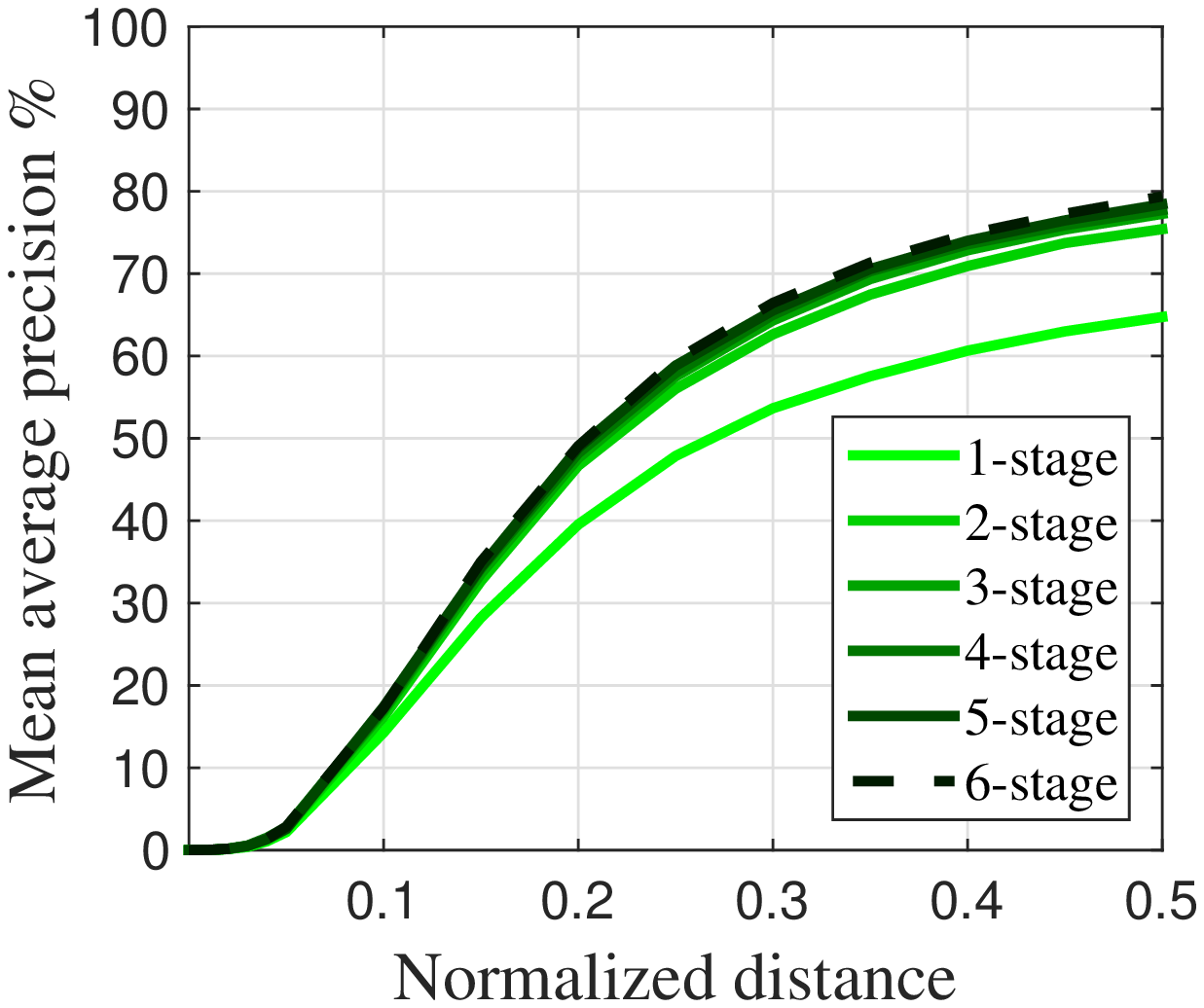}\\
    \vspace{-2pt}
  \hspace{15pt}{\footnotesize (a)}\hspace{110pt}  {\footnotesize (b) }\\
   \vspace{-5pt}
  \caption{mAP curves over different PCKh thresholds on MPII validation set. (a) mAP curves of self-comparison experiments. 
  (b) mAP curves of PAFs across stages.}
  \label{fig:mpii_curves}
\end{figure}

Fig.~\ref{fig:mpii_curves}a shows an ablation analysis on our validation set. For the threshold of PCKh-0.5~\cite{andriluka20142d},
the accuracy of our PAF method is 2.9\% higher than one-midpoint and 2.3\% higher than two intermediate points, generally outperforming the method of midpoint representation.
The PAFs, which encode both position and orientation information of human limbs, are better able to distinguish the common cross-over cases, e.g., overlapping arms. Training with masks of unlabeled persons further improves the performance by $2.3\%$ because it avoids penalizing the true positive prediction in the loss during training. If we use the ground-truth keypoint location with our parsing algorithm, we can obtain a mAP of $88.3\%$. In Fig.~\ref{fig:mpii_curves}a, the mAP obtained using our parsing with GT detection is constant across different PCKh thresholds due to no localization error. Using GT connection with our keypoint detection achieves a mAP of $81.6\%$. It is notable that our parsing algorithm based on PAFs achieves a similar mAP as when based on GT connections ($79.4\%$ vs $81.6\%$). This indicates parsing based on PAFs is quite robust in associating correct part detections. 
Fig.~\ref{fig:mpii_curves}b shows a comparison of performance across stages. The mAP increases monotonically with the iterative refinement framework. Fig.~\ref{fig:stages} shows the qualitative improvement of the predictions over stages.


\subsection{Results on the COCO Keypoints Challenge}\label{sec:coco}

The COCO training set consists of over 100K person instances labeled with over 1 million
keypoints. 
The testing set contains ``test-challenge'' and ``test-dev''subsets, which have roughly 20K images each. The COCO evaluation defines the object keypoint similarity (OKS) and uses the mean average precision (AP) over 10 OKS thresholds as the main competition metric~\cite{COCOkeypoint}. The OKS plays the same role as the IoU in object detection. It is calculated from the scale of the person and the distance between predicted and GT points. Table~\ref{table:coco} shows results from top teams in the challenge.
%
It is noteworthy that our method has a higher drop in accuracy when considering only people of higher scales ($\text{AP}^L$).


\begin{table}[t]
    \footnotesize
    \begin{center}
    \begin{tabular}{l |c| c c c c }
    \hline
    Team & \textbf{AP} & AP$^{50}$ &AP$^{75}$ &AP$^{M}$	&AP$^{L}$ \\
    \hline
    \multicolumn{6}{ c }{Top-Down Approaches} \\
    Megvii~\cite{chen2017cascaded}	&78.1	&94.1	&85.9	&74.5	&83.3	\\
    MRSA~\cite{xiao2018simple} &76.5	&92.4	&84.0	&73.0	&82.7	\\
    The Sea Monsters* &75.9 &92.1	&83.0	&71.7	&82.1 \\
    Alpha-Pose~\cite{fang2017rmpe}&	71.0&	87.9&	77.7&	69.0&	75.2	\\
    Mask R-CNN~\cite{he2017mask}&	69.2&	90.4&	76.0&	64.9&	76.3\\
    \hline
    \multicolumn{6}{ c }{Bottom-Up Approaches} \\
    METU~\cite{kocabas18prn} &70.5	&87.7	&77.2	&66.1	&77.3 \\
    TFMAN*	&70.2	&89.2	&77.0	&65.6	&76.3 \\	
    PersonLab~\cite{papandreou2018personlab} & 68.7 & 89.0 & 75.4 & 64.1 & 75.5 \\
    Associative Emb.~\cite{newell2017associative} & 65.5 & 86.8 & 72.3 & 60.6 & 72.6\\
    Ours & 64.2 & 86.2 & 70.1 & 61.0 & 68.8 \\ 
    Ours~\cite{cao2017realtime} & 61.8 & 84.9 & 67.5 & 57.1 & 68.2 \\
    \hline
    \end{tabular}
    \end{center}
    \vspace{-5pt}
    \caption{COCO test-dev leaderboard~\cite{COCOLeaderboard}, ``*'' indicates that no citation 
    was provided. Top: some of the highest top-down results. Bottom: highest bottom-up results. 
    }
    \label{table:coco}
    \vspace{-10pt}
\end{table}

\begin{table}[h]
    \footnotesize
    \begin{center}
    \begin{tabular}{l |c |c c c c }
    \hline
    Method & \textbf{AP} &	AP$^{50}$ &AP$^{75}$ &AP$^{M}$  &AP$^{L}$ \\
    \hline
    {\footnotesize GT Bbox + CPM~\cite{wei2016convolutional}} & 62.7 & \textbf{86.0} & 69.3 & 58.5 & 70.6 \\
    {\footnotesize SSD~\cite{liu2015ssd} + CPM~\cite{wei2016convolutional}} &52.7 &71.1 &57.2 &47.0 &64.2 \\
    {\footnotesize Ours~\cite{cao2017realtime}} & 58.4 & 81.5 & 62.6 & 54.4 & 65.1\\
    {\footnotesize \quad + CPM refinement} & 61.0 & 84.9 & 67.5 & 56.3 & 69.3\\
    {\footnotesize Ours} & \textbf{65.3} & 85.2 & \textbf{71.3} & \textbf{62.2} & \textbf{70.7} \\ 
    \hline
    \end{tabular}
    \end{center}
    \vspace{-10pt}
    \caption{Self-comparison experiments on the COCO validation set.
    Our new body+foot model outperforms the original work in~\cite{cao2017realtime} by 6.9\%.}
    \vspace{-3pt}
    \label{table:coco_val}
\end{table}

In Table~\ref{table:coco_val}, we report self-comparisons on 
the COCO validation set. 
If we use the GT bounding box and a single person CPM~\cite{wei2016convolutional}, we can achieve an upper-bound for the top-down approach using CPM, which is $62.7\%$ AP. If we use the state-of-the-art object detector, Single Shot MultiBox Detector (SSD)\cite{liu2015ssd}, the performance drops $10\%$. 
%
This comparison indicates the performance of top-down approaches rely heavily on the person detector. In contrast, our original bottom-up method achieves $58.4\%$ AP. If we refine the results 
by applying a single person CPM on each rescaled region of the estimated persons parsed by our method, we gain a $2.6\%$ overall AP increase.
We only update estimations on predictions
in which both methods roughly agree,
resulting in improved precision and recall.
The new architecture without CPM refinement is approximately 7\% more accurate than the original approach, while increasing the speed $\times$2.

\begin{table}[h]
\footnotesize
\begin{center}
\begin{tabular}{l |c |c c c c | c}
\hline
Method & \textbf{AP} &	AP$^{50}$ &AP$^{75}$ &AP$^{M}$  &AP$^{L}$ & Stages\\
\hline
{\footnotesize 5 PAF - 1 CM} & \textbf{65.3} & 85.2 & 71.3 & 62.2 & \textbf{70.7} & 6\\
{\footnotesize 4 PAF - 2 CM} & 65.2 & \textbf{85.3} & \textbf{71.4} & 62.3 & 70.1 & 6 \\
{\footnotesize 3 PAF - 3 CM} & 65.0 & 85.1 & 71.2 & \textbf{62.4} & 69.4 & 6 \\
{\footnotesize 4 PAF - 1 CM} & 64.8 & \textbf{85.3} & 70.9 & 61.9 & 69.6 & 5 \\
{\footnotesize 3 PAF - 1 CM} & 64.6 & 84.8 & 70.6 & 61.8 & 69.5 & 4 \\
{\footnotesize 3 CM - 3 PAF} & 61.0 & 83.9 & 65.7 & 58.5 & 65.3 & 6 \\
\hline
\end{tabular}
\end{center}
\vspace{-10pt}
\caption{Self-comparison experiments on the COCO validation set. CM refers to confidence map, while the numbers express the number of estimation stages for PAF and CM. \textit{Stages} refers to the number of PAF and CM stages. Reducing the number of stages increases the runtime performance.}
\label{table:number_paf_hm}
\end{table}

We analyze the effect of PAF refinement over confidence map estimation in Table~\ref{table:number_paf_hm}. We fix the computation to a maximum of 6 stages, distributed differently across the PAF and confidence map branches. We can extract 3 conclusions from this experiment. First, PAF requires a higher number of stages to converge and benefits more from refinement stages. Second, increasing the number of PAF channels mainly improves the number of true positives, even though they might not be too accurate (higher $AP^{50}$). However, increasing the number of confidence map channels further improves the localization accuracy (higher $AP^{75}$). Third, we prove that the accuracy of the part confidence maps highly increases when using PAF as a prior, while the opposite results in a 4\% absolute accuracy decrease. Even the model with only 4 stages (3 PAF - 1 CM) is more accurate than the computationally more expensive 6-stage model that first predicts confidence maps (3 CM - 3 PAF).
Some other additions that further increased the accuracy of the new models with respect to the original work are PReLU over ReLU layers and Adam optimization instead of SGD with momentum.
Differently to~\cite{cao2017realtime}, we do not refine the current approach with CPM~\cite{wei2016convolutional} to avoid harming the speed.

\begin{figure}[t]
\centering
    \includegraphics[width=0.7\linewidth]{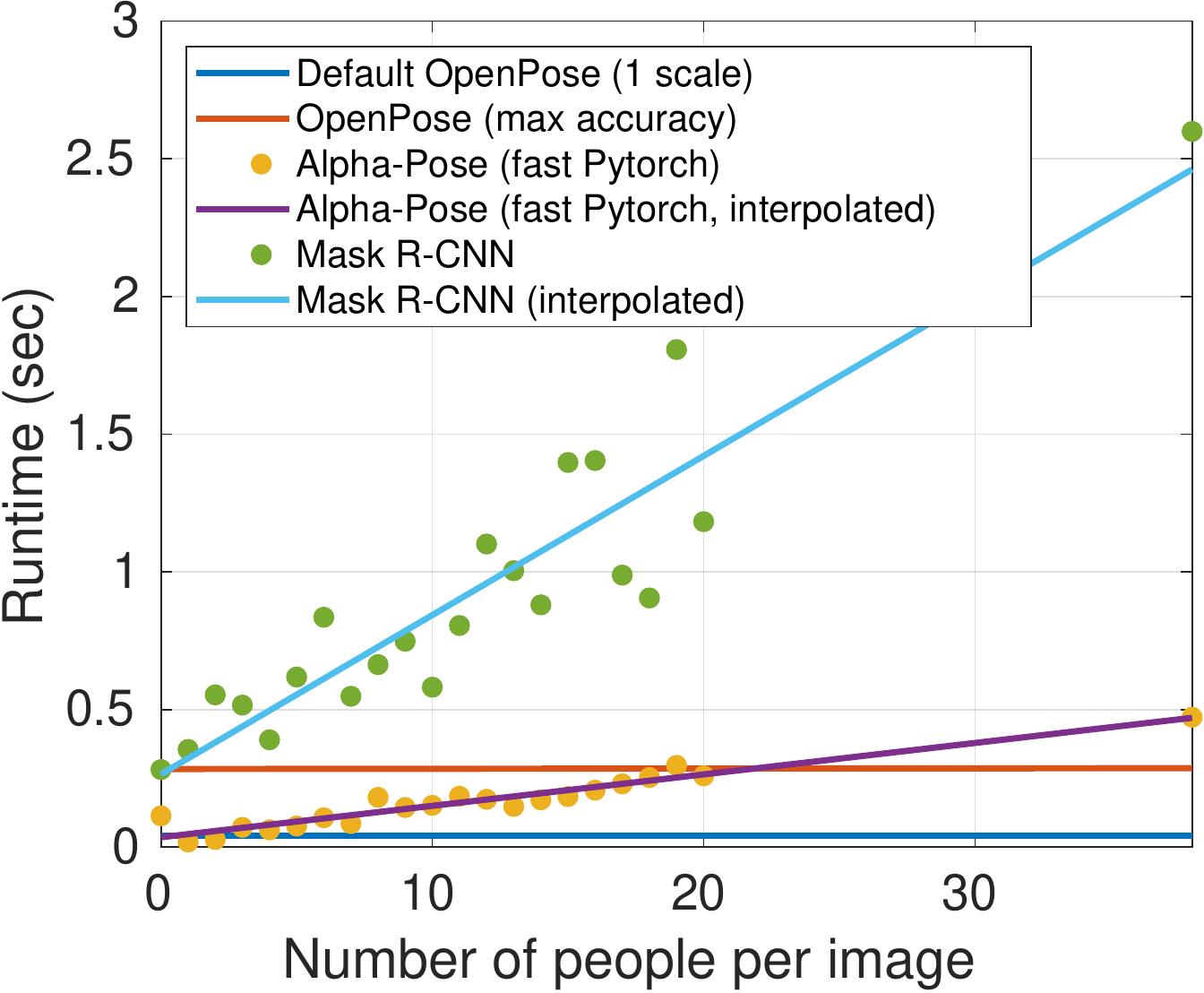}
    \vspace{-5pt}
    \caption{Inference time comparison between OpenPose, Mask R-CNN, and Alpha-Pose (fast Pytorch version). While OpenPose inference time is invariant, Mask R-CNN and Alpha-Pose runtimes grow linearly with the number of people.
    Testing with and without scale search is denoted as ``max accuracy" and ``1 scale", respectively. This analysis was performed using the same images for each algorithm and a batch size of 1. Each analysis was repeated 1000 times and then averaged. This was all performed on a system with a Nvidia 1080 Ti and CUDA 8.}
    \vspace{-10pt}
\label{fig:op_vs_mrcnn}
\end{figure}

\subsection{Inference Runtime Analysis}\label{sec:runtime}
We compare
3 state-of-the-art, well-maintained, and widely-used multi-person pose estimation libraries, OpenPose~\cite{hidalgo_cao_simon_wei_joo_sheikh_2017}, based on this work, Mask R-CNN~\cite{he2017mask}, and Alpha-Pose~\cite{fang2017rmpe}. We analyze the inference runtime performance of the 3 methods in Fig.~\ref{fig:op_vs_mrcnn}. Megvii (Face++)~\cite{chen2017cascaded} and MSRA~\cite{xiao2018simple} GitHub repositories do not include the person detector they use and only provide pose estimation results given a cropped person. Thus, we cannot know their exact runtime performance and have been excluded from this analysis.
Mask R-CNN is only compatible with Nvidia graphics cards, so we perform the analysis on a
system with a NVIDIA 1080 Ti.
As top-down approaches, the inference times of Mask R-CNN, Alpha-Pose, Megvii, and MSRA are roughly proportional to the number of people in the image. To be more precise, they are proportional to the number of
proposals that their person detectors extract. In contrast, the inference time of our bottom-up approach is invariant to the number of people in the image.
The runtime of OpenPose consists of two major parts: (1) CNN processing time whose 
complexity is $O(1)$, constant with varying number of people; (2) multi-person parsing time, whose 
complexity is $O(n^2)$, where $n$ represents the number of people. However, the parsing time 
is two orders of magnitude less than the CNN processing time. For instance, the parsing takes $0.58$ ms for $9$ people while the CNN takes $36$ ms. 

\begin{table}[h]
    \footnotesize
    \begin{center}
    \begin{tabular}{l |c c c }
    \hline
    Method & CUDA & CPU-only \\
    \hline
    {\footnotesize Original MPII model} & 73 ms & \textbf{2309 ms}  \\ 
    {\footnotesize Original COCO model} & 74 ms & 2407 ms  \\ 
    {\footnotesize Body+foot model}     & \textbf{36 ms} & 10396 ms \\ 
    \hline
    \end{tabular}
    \end{center}
    \vspace{-7pt}
    \caption{Runtime 
    difference between the 3 models released in OpenPose with CUDA and CPU-only versions, running
    in a NVIDIA GeForce GTX-1080 Ti GPU and a i7-6850K CPU.
    MPII and COCO models refer to our work in \cite{cao2017realtime}.}
    \label{table:inference_time}
    \vspace{-2pt}
\end{table}

In Table~\ref{table:inference_time}, we analyze the difference in inference time between the models released in OpenPose, i.e., the MPII and COCO models from \cite{cao2017realtime} and the new body+foot model. Our new combined model is not only more accurate, but
is also $\times$2 faster than the original model when using the GPU version.
Interestingly, the runtime for the CPU version is 5x slower compared to that of the original model. The new architecture consists of many more layers, which requires a
higher amount of memory,
while the number of operations is significantly fewer. Graphic cards seem to benefit more from the reduction in number of operations, while the CPU version seems to be significantly slower due to the higher memory requirements. OpenCL and CUDA performance cannot be directly compared to each other, as they require different hardware, in particular, different GPU brands.


\subsection{Trade-off between Speed and Accuracy}




In the area of object detection, Huang \textit{et al.}~\cite{huang2017speed} show that region-proposed methods (e.g., Faster-rcnn~\cite{ren2015faster}) achieve higher accuracy, while single-shot methods (e.g., YOLO~\cite{redmon2017yolo9000}, SSD~\cite{liu2015ssd}) present higher runtime performance.
Analogously in human pose estimation, we observe that top-down approaches also present higher accuracy but lower speed compared to bottom-up methods, especially for images with multiple people.
The main reason for the lower accuracy of bottom-up approaches is
their limited resolution.
While top-down methods individually crop and feed each detected person into their networks, bottom-up methods have to feed the whole image at once, resulting in smaller resolution per person. For instance, Moon \textit{et al.}~\cite{moon2018posefix} show that refinement over our original work in~\cite{cao2017realtime} (by applying a larger cropped image patch) results in a higher accuracy boost than refinement over other top-down approaches. As hardware gets faster and increases its memory, bottom-up methods with higher resolution might be able to reduce the accuracy gap with respect to top-down approaches. 

\begin{figure}[t]
\centering
    \includegraphics[width=0.4875\linewidth]{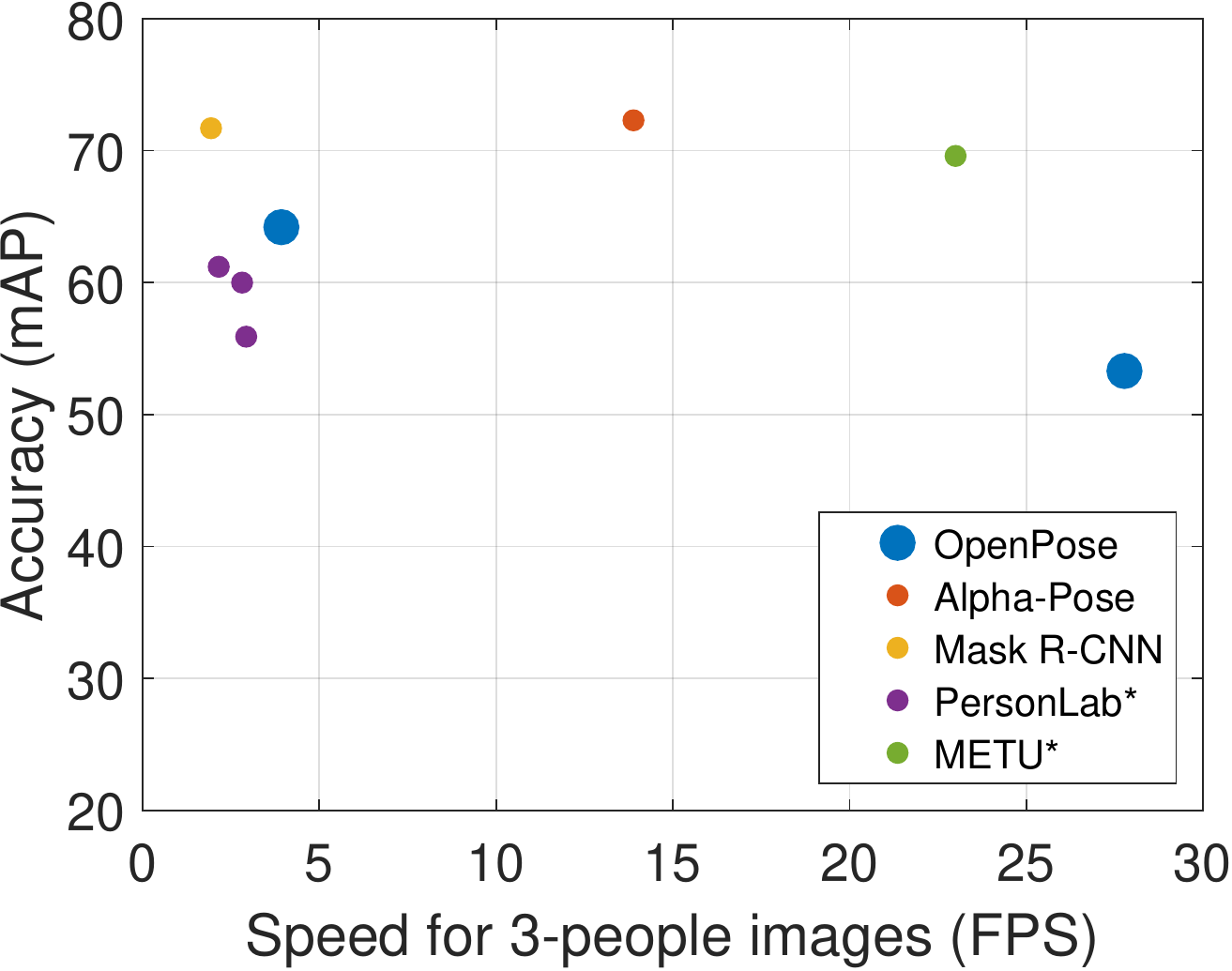}
    \includegraphics[width=0.4875\linewidth]{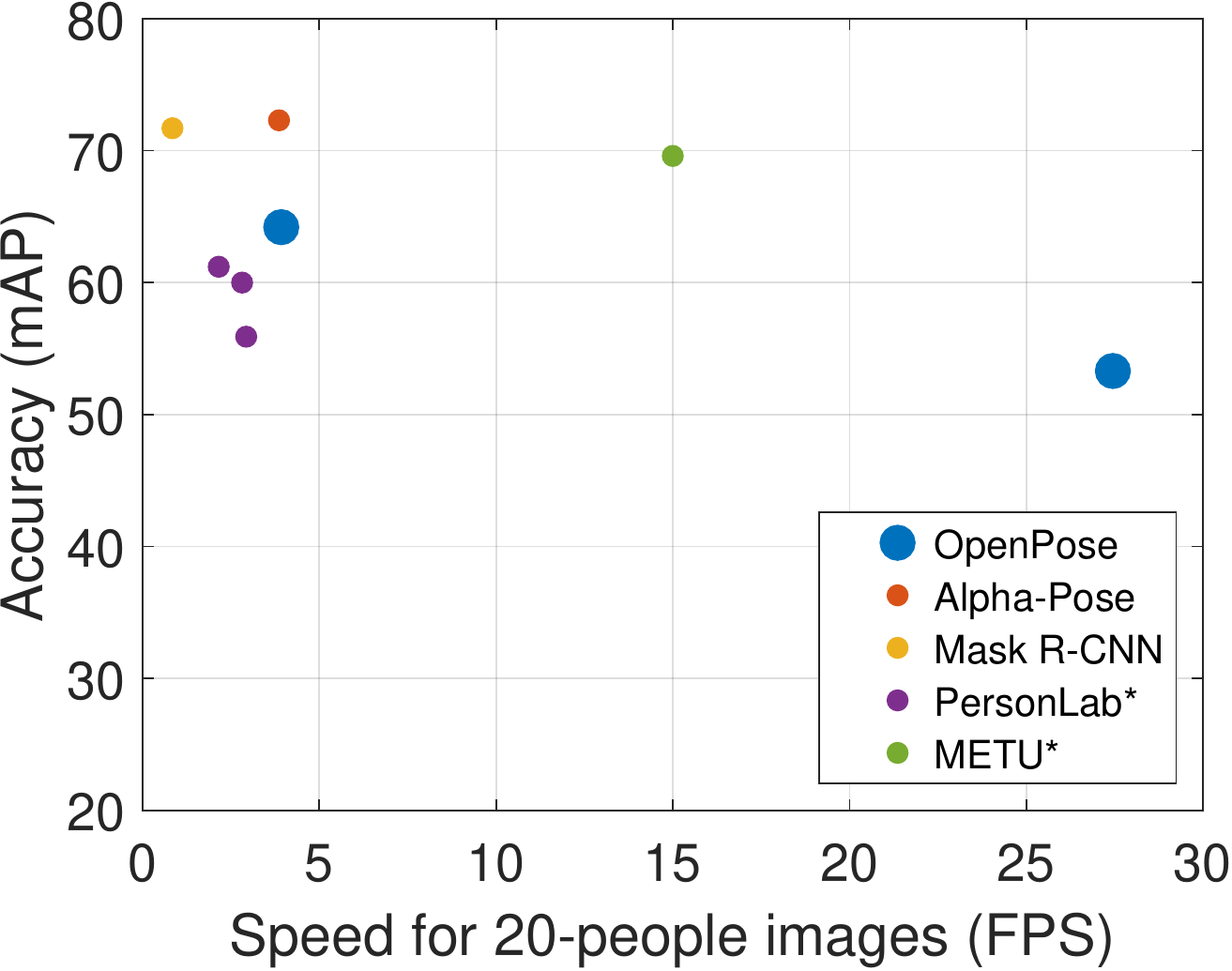}
    \vspace{-2pt}
    \caption{Trade-off between speed and accuracy for the main entries of the COCO Challenge. We only consider those approaches that either release their runtime measurements (methods with an asterisk) or their code (rest). Algorithms with several values represent different resolution configurations. AlphaPose, METU, and single-scale OpenPose provide the best results considering the trade-off between speed and accuracy. The remaining methods are both slower and less accurate than at least one of these 3 approaches.}
    \vspace{-10pt}
\label{fig:trade_off_speed_accuracy}
\end{figure}
Additionally, current human pose performance metrics are purely based on keypoint accuracy, while speed is ignored. 
In order to provide a more complete comparison, we display both speed and accuracy for the top entries of the COCO Challenge in Fig.~\ref{fig:trade_off_speed_accuracy}.
Given those results, single-scale OpenPose should be chosen for maximum speed, AlphaPose for maximum accuracy, and METU for a trade-off between both of them. The remaining approaches are both slower and less accurate than at least one of these 3 methods.
Overall, top-down approaches (e.g., AlphaPose) provide better results for images with few people, but their speed considerably drops for images with many people.
We also observe that the accuracy metrics might be misleading. We see in Section~\ref{sec:coco} 
that PersonLab~\cite{papandreou2018personlab} achieves higher accuracy than our method. However, our multi-scale approach simultaneously provides both higher speed and accuracy than the versions for which they report runtime results. Note that no runtime results are provided in~\cite{papandreou2018personlab} for their most accurate (but slower) models.

\begin{figure*}[h!]
    \centering
    \begin{subfigure}[t]{0.3175\linewidth} 
        \centering
        \includegraphics[width=1\linewidth]{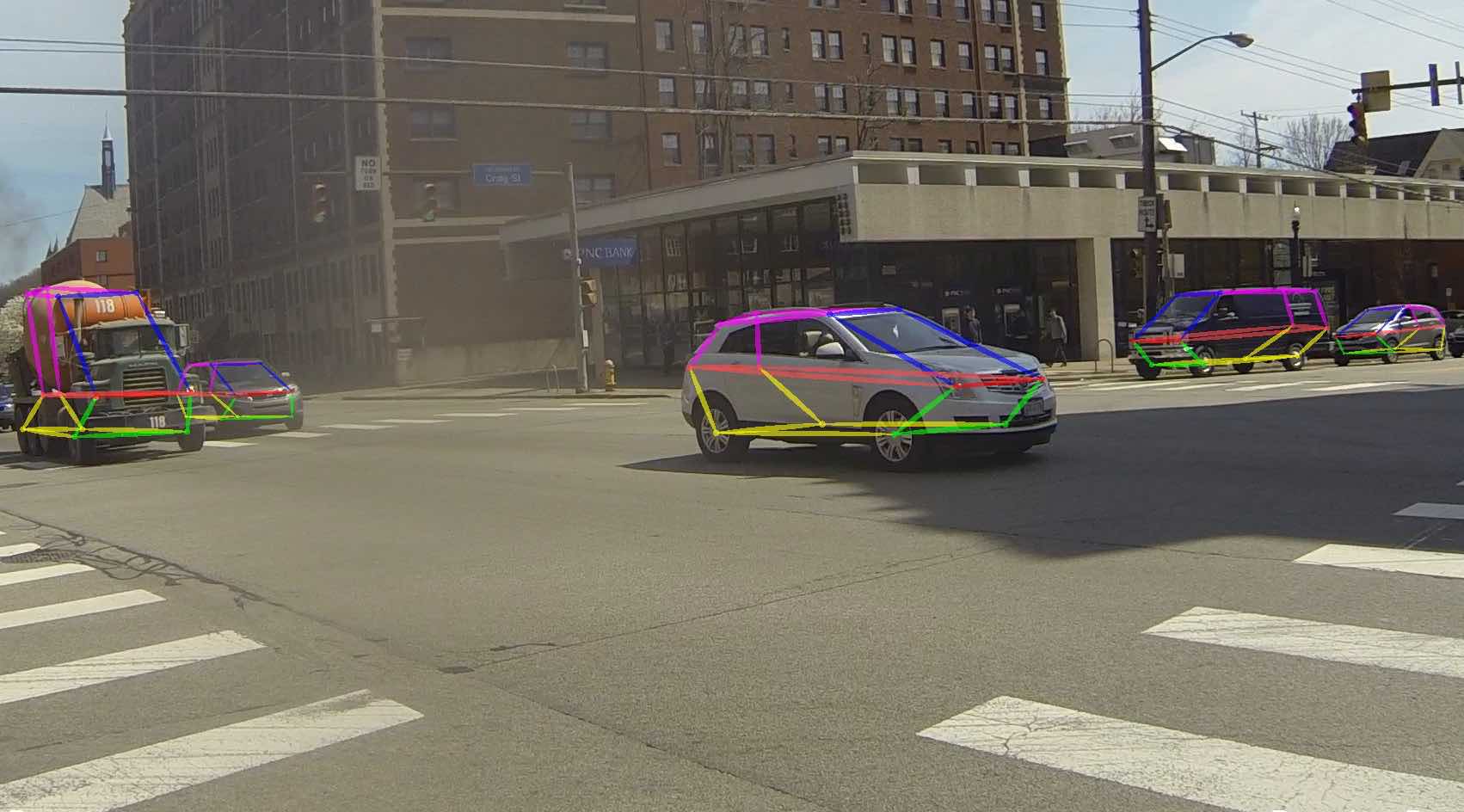}
    \end{subfigure}
    \begin{subfigure}[t]{0.3175\linewidth} 
        \centering
        \includegraphics[width=1\linewidth]{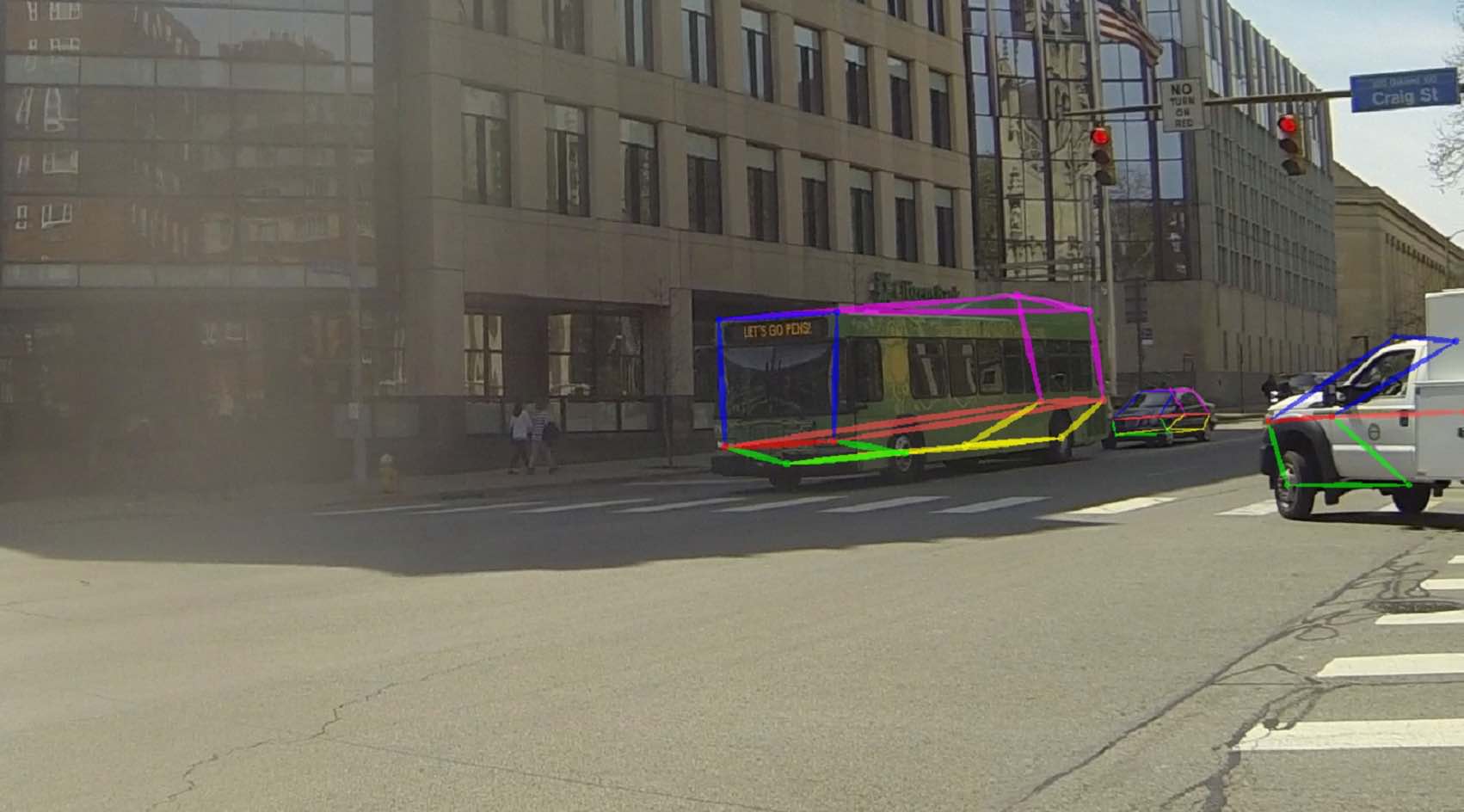}
    \end{subfigure}
    \begin{subfigure}[t]{0.3175\linewidth} 
        \centering
        \includegraphics[width=1\linewidth]{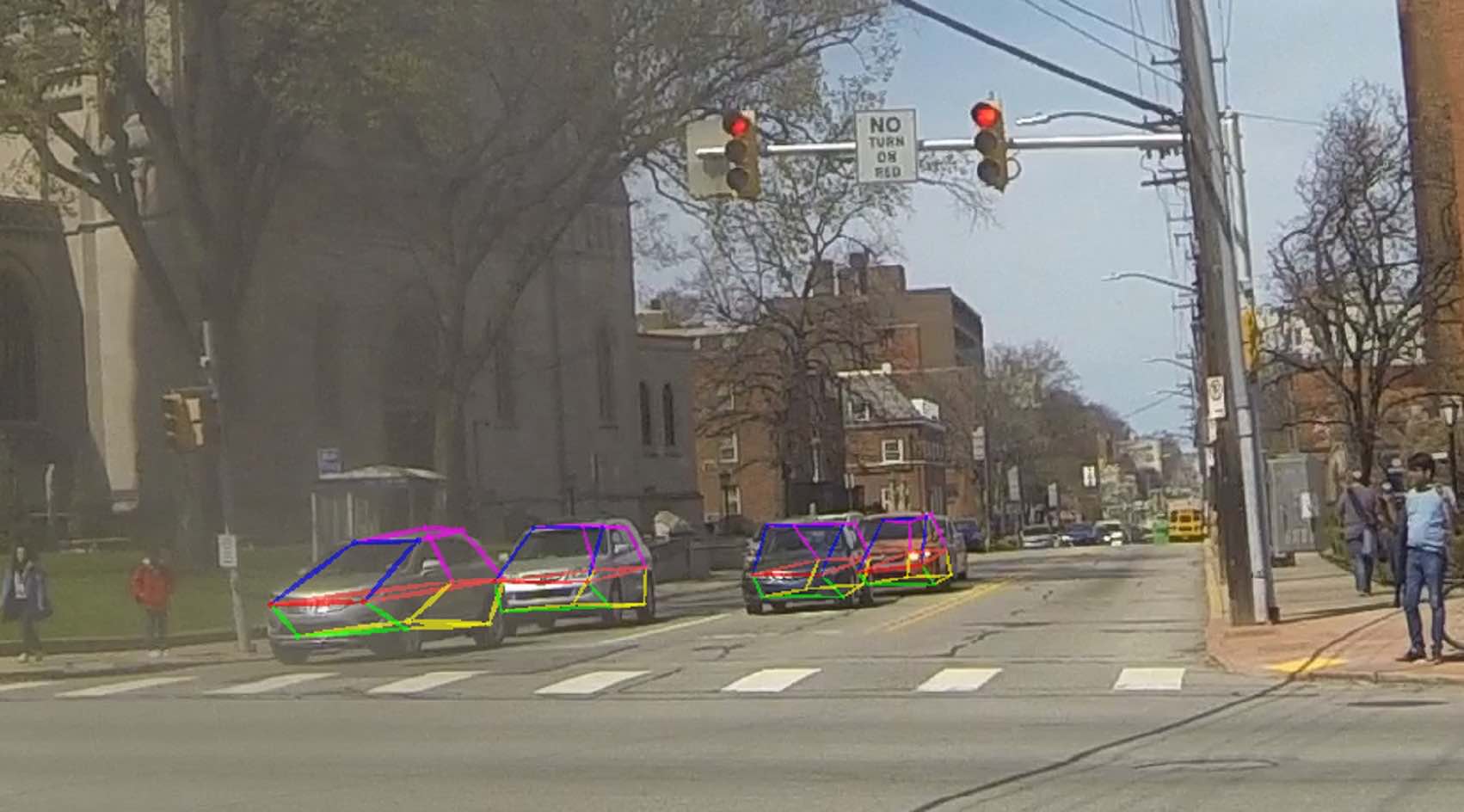}
    \end{subfigure}
    \vspace{-4pt}
    \caption{Vehicle keypoint detection examples from the validation set. The keypoint locations are successfully estimated under challenging scenarios, including overlapping between cars, cropped vehicles, and different scales.}
    \label{fig:vehicle}
    \vspace{-4pt}
\end{figure*}

\begin{figure*}[t]
\begin{center}
\includegraphics[width=0.96\linewidth]{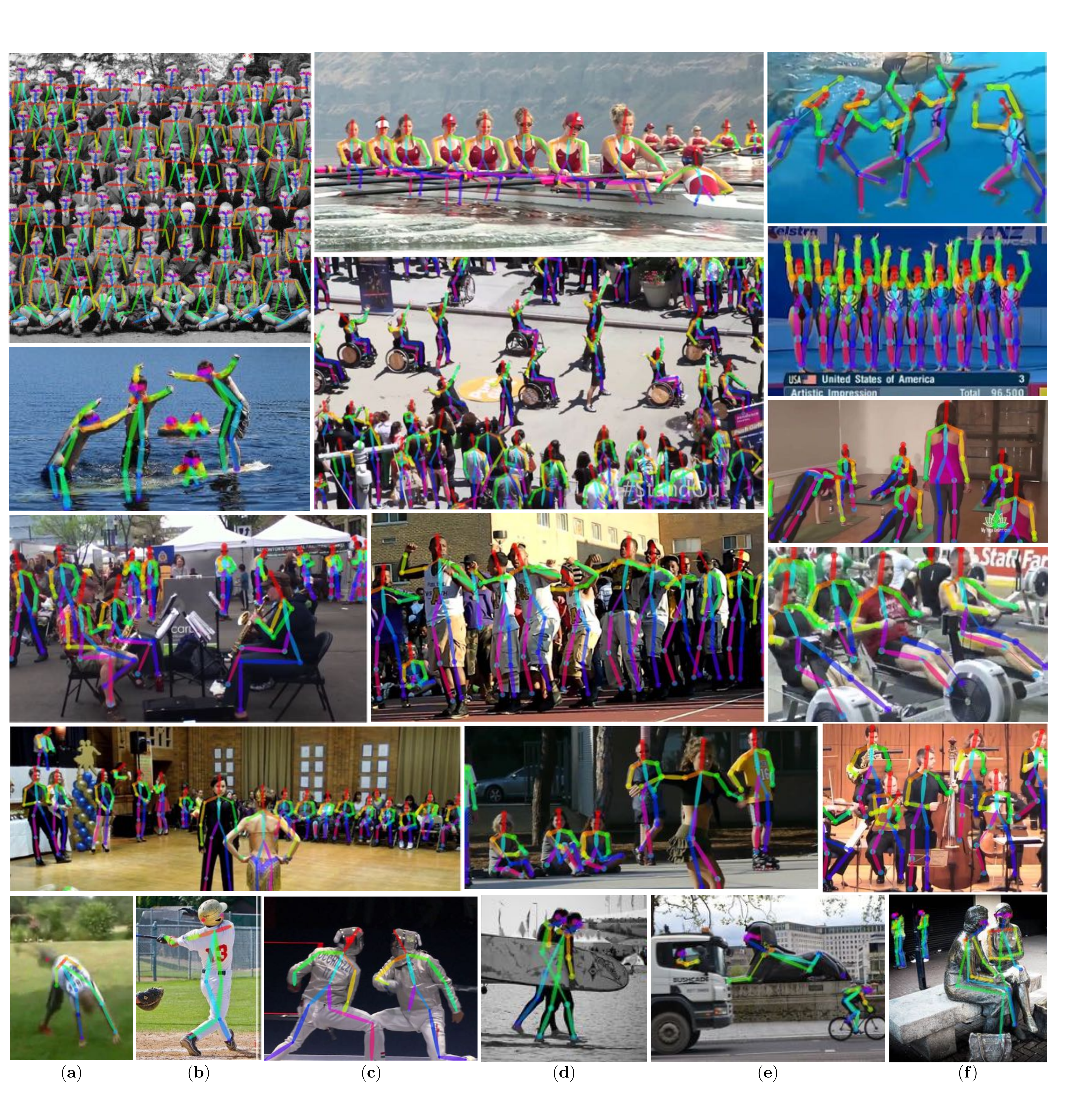}
\end{center}
\vspace{-12pt}
   \caption{Common failure cases: (a) rare pose or appearance, (b) missing or false parts detection, (c) overlapping parts, i.e., part detections shared by two persons, (d) wrong connection associating parts from two persons, (e-f): false positives on statues or animals.}
\vspace{-4pt}
\label{fig:failure}
\end{figure*}

\begin{figure*}[t]
\centering
    \begin{subfigure}[t]{0.1325\textwidth}
        \centering
        \includegraphics[width=1\linewidth]{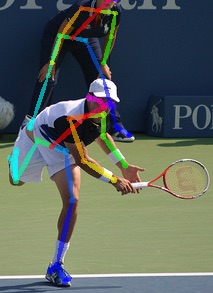}
        \vspace{-16pt}
        \caption{}
    \end{subfigure}
    \begin{subfigure}[t]{0.1325\textwidth}
        \centering
        \includegraphics[width=1\linewidth]{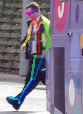}
        \vspace{-16pt}
        \caption{}
    \end{subfigure}
    \begin{subfigure}[t]{0.1325\textwidth}
        \centering
        \includegraphics[width=1\linewidth]{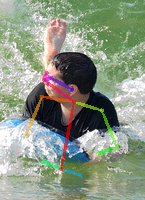}
        \vspace{-16pt}
        \caption{}
    \end{subfigure}
    \begin{subfigure}[t]{0.1325\textwidth}
        \centering
        \includegraphics[width=1\linewidth]{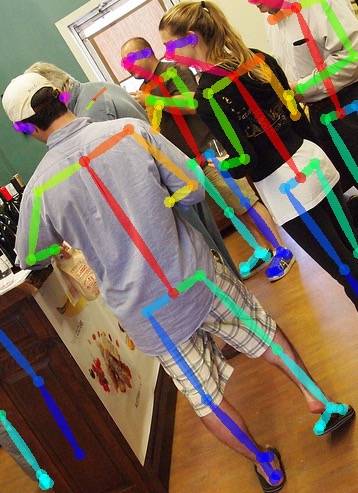}
        \vspace{-16pt}
        \caption{}
    \end{subfigure}
    \begin{subfigure}[t]{0.1325\textwidth}
        \centering
        \includegraphics[width=1\linewidth]{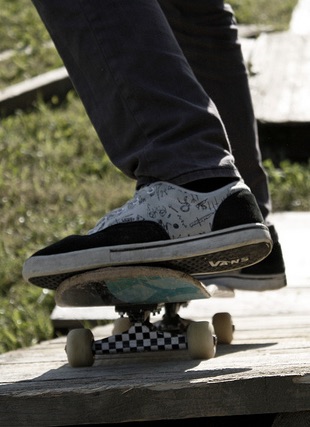}
        \vspace{-16pt}
        \caption{}
    \end{subfigure}
    \begin{subfigure}[t]{0.1325\textwidth}
        \centering
        \includegraphics[width=1\linewidth]{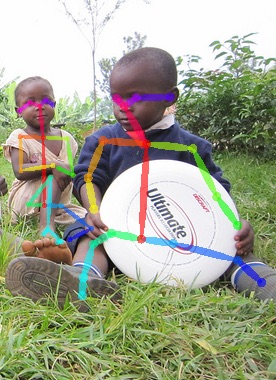}
        \vspace{-16pt}
        \caption{}
    \end{subfigure}
    \begin{subfigure}[t]{0.1325\textwidth}
        \centering
        \includegraphics[width=1\linewidth]{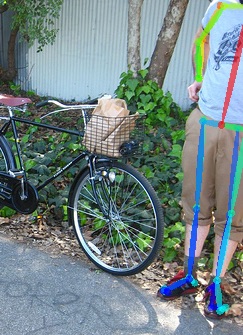}
        \vspace{-16pt}
        \caption{}
    \end{subfigure}
    \vspace{-8pt}
   \caption{Common foot failure cases: (a) foot or leg occluded by the body, (b) foot or leg occluded by another object, (c) foot visible but leg occluded, (d) shoe and foot not aligned, (e): false negatives when foot visible but rest of the body occluded, (f): soles of their feet are usually not detected (rare in training), (g): swap between right and left body parts.}
    \vspace{-7pt}
\label{fig:failure_foot}
\end{figure*}

\subsection{Results on the Foot Keypoint Dataset}\label{sec:results_foot}
To evaluate the foot keypoint detection results obtained using our foot keypoint dataset, we calculate the mean average precision and recall over 10 OKS, as done in the COCO
evaluation metric. There are only minor differences between the combined and body-only approaches. In the combined training scheme, there exist two separate and completely independent datasets. The larger of the two datasets consists of the body annotations while the smaller set contains both body and foot annotations. The same batch size used for the body-only training is used for the combined training. Nevertheless, it contains only annotations from one dataset at a time. A probability ratio is defined to select the dataset from which to pick each batch. A higher probability is assigned to select a batch from the larger dataset, as the number of annotations and diversity is much higher. Foot keypoints are masked out during the back-propagation pass of the body-only dataset to avoid harming the net with non-labeled data. In addition, body annotations are also masked out from the foot dataset.
Keeping these annotations yields a small drop in accuracy,
probably due to overfitting, as those samples are repeated in both datasets.

\begin{table}[h]
    \footnotesize
    \begin{center}
    \begin{tabular}{l |c c | c c}
    \hline
    Method & \textbf{AP} & \textbf{AR} & AP$^{75}$ &AR$^{75}$ \\
    \hline
    {\footnotesize Body+foot model (5 PAF - 1 CM)} & 77.9 & 82.5 & 82.1 & 85.6 \\
    \hline
    \end{tabular}
    \end{center}
    \vspace{-10pt}
    \caption{Foot keypoint analysis on the foot validation set.}
    \vspace{-2pt}
    \label{table:foot_whole_vs_mixed}
\end{table}

Table~\ref{table:foot_whole_vs_mixed} shows the foot keypoint accuracy for our validation set. This set is created from a subset of the COCO validation set, in particular from the images in which the ankles of all people are visible and annotated. This results in a simpler validation set compared to that of COCO, leading to higher precision and recall numbers compared to those of body detection (Table~\ref{table:coco_val}).
Qualitatively, we find a higher amount of jitter and number of detection errors compared to body keypoint prediction. We believe 14K training annotations are not a sufficient number to train a robust foot detector, considering that over 100K instances are used for the body keypoint dataset. 
Rather than using the whole batch with either only foot or body annotations, we also tried using a mixed batch where samples from both datasets (either COCO or COCO+foot) could be fed to the same batch, maintaining the same probability ratio. However, the network accuracy
was slightly reduced.
By mixing the datasets with an unbalanced ratio, we effectively assign a very small batch size for foot, hindering foot convergence. 

\begin{table}[h]
\vspace{+2pt}
\footnotesize
\begin{center}
\begin{tabular}{l |c |c c c c }
\hline
Method & \textbf{AP} &	AP$^{50}$ &AP$^{75}$ &AP$^{M}$  &AP$^{L}$ \\
\hline
{\footnotesize Body-only (5 PAF - 1 CM)} & 65.2 & 85.0 & 70.9 & 62.1 & 70.5 \\
{\footnotesize Body+foot (5 PAF - 1 CM)} & \textbf{65.3} & \textbf{85.2} & \textbf{71.3} & \textbf{62.2} & \textbf{70.7} \\
\hline
\end{tabular}
\end{center}
\vspace{-7pt}
\caption{Self-comparison experiments for body on the COCO validation set. Foot keypoints are predicted but ignored for the evaluation.}
\label{table:body_vs_foot}
\end{table}

In Table~\ref{table:body_vs_foot}, we show
that there is almost no accuracy difference
in the COCO test-dev set with respect to the same network architecture trained only with body annotations. We compared the model consisting of 5 PAF and 1 confidence map stages, with a 95\% probability of picking a batch from the COCO body-only dataset, and 5\% 
of choosing from the body+foot dataset. There is no architecture difference compared to the body-only model other than the increase in the number of outputs to include the foot CM and PAFs.

\subsection{Vehicle Pose Estimation}\label{sec:vehicle}
Our approach is not limited to human body or foot keypoints, but can be generalized to any keypoint annotation task. To demonstrate this, we have run the
same network architecture for the task of vehicle keypoint detection~\cite{dinesh2018carfusion}. Once again, we use mean average precision over 10 OKS for the evaluation. The results are shown in Table~\ref{table:vehicle}. Both the average precision and recall are higher than in the body keypoint task, mainly because we are using a smaller and simpler dataset. This initial dataset consists of image annotations from 19 different cameras. We have used the first 18 camera frames as a training set, and the camera frames from the last camera as a validation set. No variations in the model architecture or training parameters have been made. We show qualitative results in Fig.~\ref{fig:vehicle}.

\begin{table}[h]
\footnotesize
\begin{center}
\begin{tabular}{l |c c | c c}
\hline
Method & \textbf{AP} & \textbf{AR} & AP$^{75}$ &AR$^{75}$ \\
\hline
{\footnotesize Vehicle keypoint detector} & 70.1 & 77.4 & 73.0 & 79.7 \\
\hline
\end{tabular}
\end{center}
\vspace{-10pt}
\caption{Vehicle keypoint validation set.}
\vspace{-12pt}
\label{table:vehicle}
\end{table}

\begin{figure*}[t]
\centering
\includegraphics[width=0.985\linewidth]{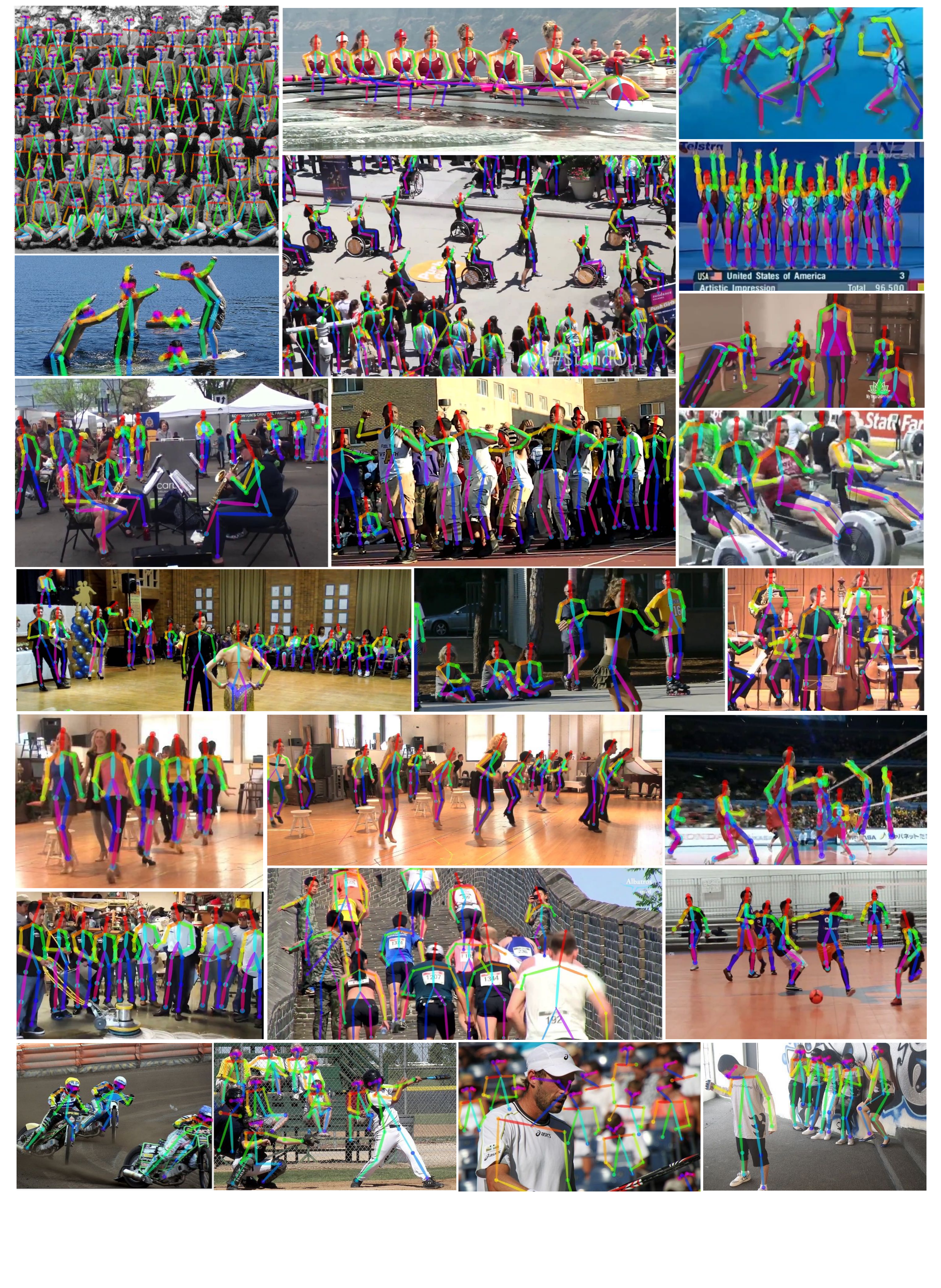}\\
\caption{Results containing viewpoint and appearance variation, occlusion, crowding, contact, and other common imaging artifacts.}
\label{fig:qua}
\end{figure*}

\subsection{Failure Case Analysis}
We have analyzed the main cases where the current approach fails in the MPII, COCO, and COCO+foot validation sets. Fig.~\ref{fig:failure} shows an overview of the main body failure cases, while Fig.~\ref{fig:failure_foot} shows the main foot failure cases Fig.~\ref{fig:failure}a refers to non typical poses and upside-down examples, where the predictions usually fail. Increasing the rotation augmentation visually seems to partially solve these issues, but the global accuracy on the COCO validation set is reduced by about 5\%. A different alternative is to run the network using different rotations and keep the poses with the higher confidence. Body occlusion can also lead to false negatives and high localization error. This problem is inherited from the dataset annotations, in which occluded keypoints are not included. In highly crowded images where people are overlapping, the approach tends to merge annotations from different people, while missing others, due to the overlapping PAFs that make the greedy multi-person parsing fail. Animals and statues also frequently lead to false positive errors. This issue could be mitigated by adding more negative examples during training to help the network distinguish between humans and other humanoid figures.
